\documentclass[10pt,twocolumn,letterpaper]{article}

\usepackage{cvpr}      
\definecolor{cvprblue}{rgb}{0.21,0.49,0.74}
\usepackage[pagebackref,breaklinks,colorlinks,allcolors=cvprblue]{hyperref}
\usepackage{xcolor} 
\usepackage{booktabs}              
\usepackage{caption}              
\usepackage{graphicx}              
\usepackage{multirow} 
\usepackage{float}
\usepackage{balance}
\usepackage{comment}
\usepackage[dvipsnames]{xcolor}
\usepackage{pifont}
\usepackage{xcolor}
\usepackage{multirow}
\usepackage{booktabs}
\usepackage{array}
\newcommand{\cmark}{\textcolor{green!60!black}{\ding{51}}}
\newcommand{\xmark}{\textcolor{red!70!black}{\ding{55}}}

\def\paperID{14524} 
\def\confName{CVPR}
\def\confYear{2026}

\title{Omni-Fake: Benchmarking Unified Multimodal Social Media \\ Deepfake Detection}

\author{
Tianxiao Li$^{1*}$ \quad
Zhenglin Huang$^{1*}$\quad
Haiquan Wen$^{1*}$ \quad \\
Yiwei He$^{1}$ \quad
Xinze Li$^{1}$ \quad
Bingyu Zhu$^{1}$ \quad
Wuhui Duan$^{1}$ \quad
Congang Chen$^{1}$ \quad \\
Zeyu Fu$^{2}$ \quad
Yi Dong$^{1}$ \quad
Baoyuan Wu$^{3}$ \quad
Jason Li$^{4}$ \quad
Guangliang Cheng$^{1}$ \\
{\small
$^{1}$University of Liverpool 
} \\
{\small
$^{2}$University of Exeter \quad
$^{3}$The Chinese University of Hong Kong, Shenzhen \quad
$^{4}$Nanyang Technological University
} \\
{\small Corresponding to: Guangliang.Cheng@liverpool.ac.uk, * means equal contribution.} \\
}

\begin{document}
\maketitle
\begin{abstract}

Multimodal deepfakes are proliferating on social media and threaten authenticity, information integrity, and digital forensics. 
Existing benchmarks are constrained by their single-modality scope, simplified manipulations, or unrealistic distributions, which limit their ability to assess real-world robustness. 
To address these limitations, we present \textbf{Omni-Fake}, a unified omni-dataset for comprehensive multimodal deepfake detection in social-media settings. 
It comprises \textbf{Omni-Fake-Set}, a large-scale, high-quality dataset with \textbf{1M+} samples, and \textbf{Omni-Fake-OOD}, an out-of-distribution benchmark with \textbf{200k+} samples intentionally excluded from training to evaluate generalization. 
Omni-Fake spans four modalities (image, audio, video, and audio-video talking head) and supports a joint detection–localization–explanation protocol.
On top of Omni-Fake, we further propose \textbf{Omni-Fake-R1}, a reinforcement-learning-driven multimodal detector that adaptively integrates visual and auditory cues and outputs structured decisions, localization, and natural-language explanations. 
Extensive experiments show significant gains in detection accuracy, cross-modal generalization, and explainability over state-of-the-art baselines. 
Project page: \url{https://tianxiao1201.github.io/omni-fake-project-page/}
\end{abstract}    
\section{Introduction}
\label{sec:intro}



\begin{figure}[t]  
  \centering
  \includegraphics[width=\linewidth]{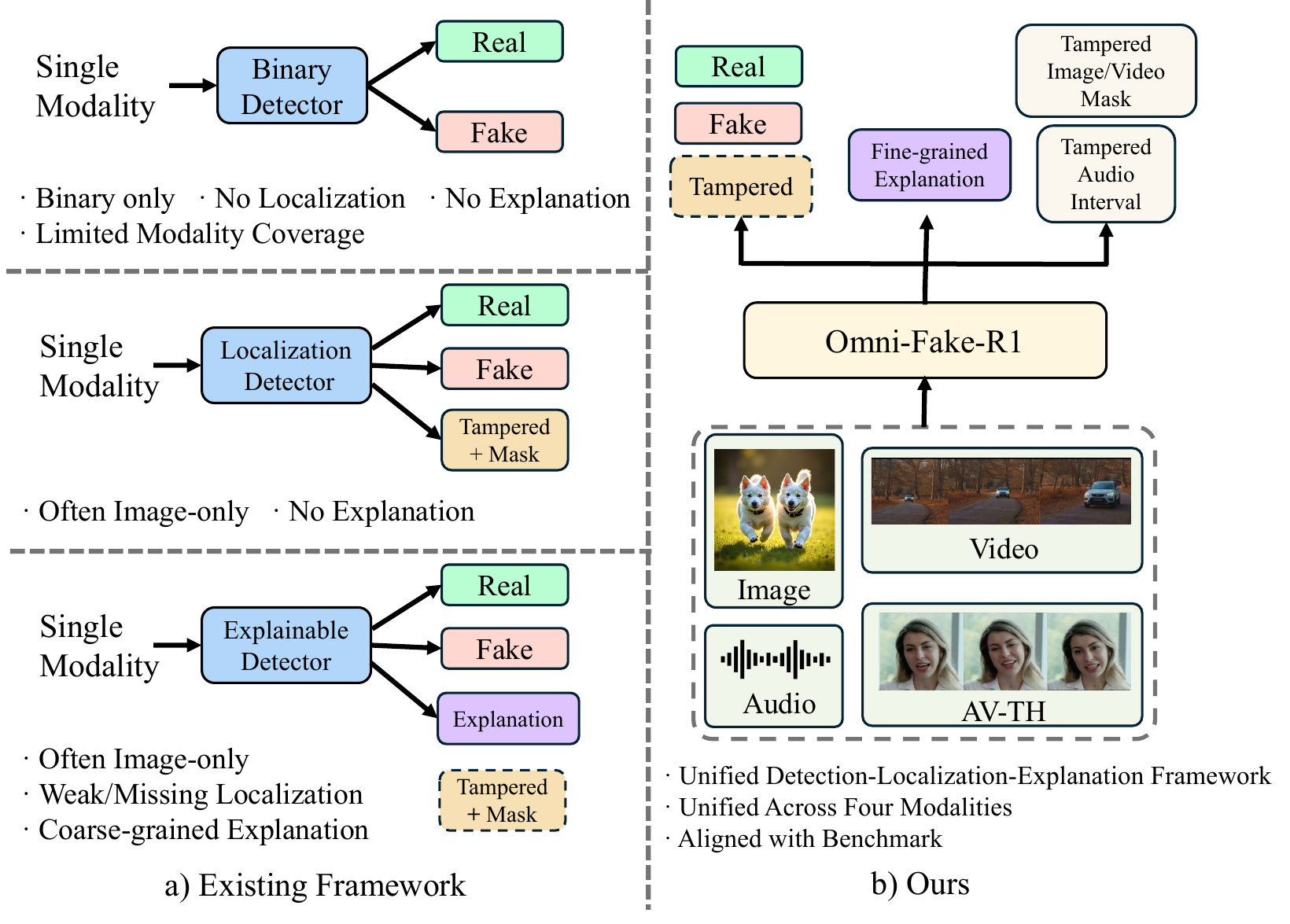}
  \caption{Framework comparison. Existing non-Omni methods like CnnSpot~\cite{wang2020cnn}, SIDA~\cite{huang2025sida}, VideoLISA~\cite{bai2024one}, and FakeSound~\cite{xie2024fakesound} usually handle only single-modality inputs. 
  In contrast, \textbf{Omni-Fake-R1}, built on a unified omni MLLM and our Omni-Fake dataset, supports four modalities, greatly expanding deepfake coverage.
  Classical non-MLLM pipelines also struggle to jointly handle detection, localization, and explanation, whereas Omni-Fake-R1 offers full support for integrated, trustworthy analysis of manipulated content.
  }
  \label{fig:teaser}
\end{figure}




The rapid progress of generative AI has flooded social platforms with highly realistic multimodal content, from images and audio to videos and talking-head avatars, sharply raising the bar for authenticity verification~\citep{comanici2025gemini,ai2025ming,li2025omniflow,hurst2024gpt}. For example, Sora, Kling, and WanX~\citep{sora-openai,kling_video_2024,wanx_video_2024} can generate near-photorealistic videos with synchronized audio. Unlike controlled academic settings, real-world timelines combine outputs from diverse proprietary generators with complex post-processing, creating severe distribution shifts that challenge current detection methods~\citep{kukanov2025klassify,hu2025simulating}. Yet the tools and benchmarks available to combat these threats have not kept pace.

%
%
\begin{figure*}[t]
  \centering
  \includegraphics[width=0.9\textwidth]{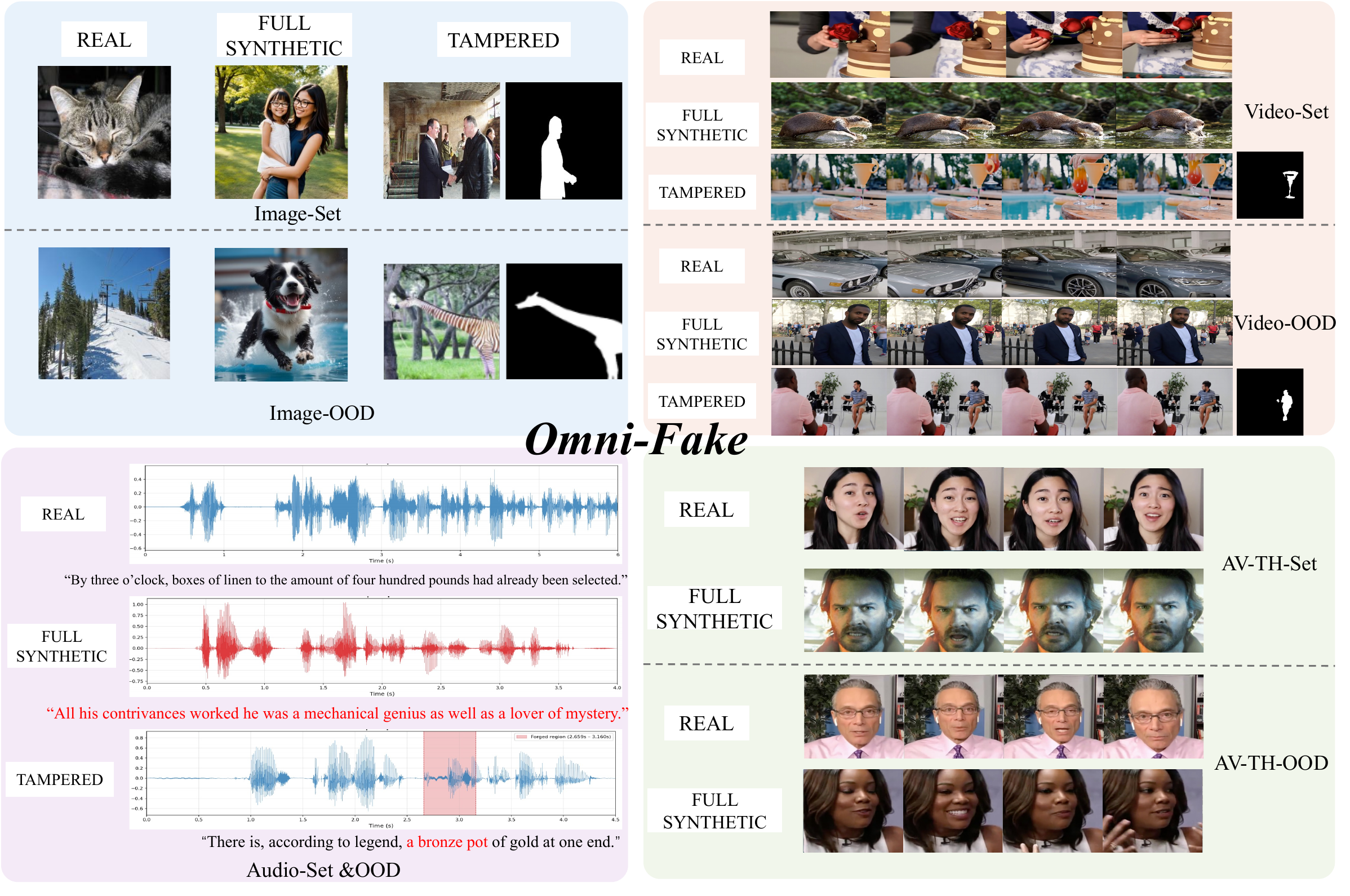}
  \caption{Representative samples from Omni-Fake across modalities, highlighting the diversity, high quality, and multimodal nature of forgeries in social media scenarios.}
  \label{fig:example}
\end{figure*}
Despite notable progress in deepfake detection~\citep{pei2024deepfake,zou2025survey,huang2025sida,xu2024fakeshield,chen2024demamba}, existing research still faces three major limitations~\citep{pei2024deepfake}. 
%
\textbf{First, benchmarks lag behind real-world practice.} Current datasets~\citep{rossler2019faceforensics,dang2020detection,he2021forgerynet,zhu2023genimage,corvi2023detection,hong2024wildfake,park2025community,huang2025so,dolhansky2020deepfake,khalid2021fakeavceleb,xu2024identity,kwon2021kodf,LOKI}  often rely on simplified generation pipelines and obsolete synthesis models, failing to systematically cover recent generators, multi-platform content formats, or multi-round adversarial attacks. Moreover, few benchmarks provide a rigorous multimodal out-of-distribution evaluation protocol or unify detection, localization, and explanation under a single assessment framework. As a result, models trained on these benchmarks tend to overfit superficial artifacts and struggle to transfer to emerging forgeries.
\textbf{Second, unified multimodal modeling remains underdeveloped.} Most detection systems~\citep{wang2020cnn,ojha2023towards,chang2023antifakeprompt,tong2022videomae,jung2022aasist,tak2022automatic,haliassos2021lips,haliassos2022leveraging,feng2023self,huang2025sida,xu2024fakeshield,chen2024demamba,li2025raidxretrievalaugmentedgenerationgrpo} are trained separately on single or paired modalities and lack a framework that jointly processes unimodal and multimodal inputs. This fragmented design leads to brittle cross-modal reasoning, weak generalization across content types, and inconsistent outputs when deployed across diverse social media environments~\citep{huang2025sida,xu2024fakeshield,chen2024demamba}.
\textbf{Third, decision processes lack transparency.} Mainstream approaches~\citep{wang2020cnn,ojha2023towards,jung2022aasist,haliassos2021lips,guo2025language,huang2025sida,bai2024one,mansoor2025explainable,zhao2025deepfakebench} default to binary classification without revealing key forged regions, cross-modal inconsistencies, or the reasoning behind the verdict. Detection, localization, and explanation are typically handled by separate modules with no consistency checks across spatial, temporal, and semantic dimensions, limiting their value for content moderation and forensic analysis~\citep{mansoor2025explainable,zhao2025deepfakebench}.

To address these limitations, we pair a comprehensive new benchmark with a unified detection model trained end-to-end for detection, localization, and explanation. Specifically, we introduce \textbf{Omni-Fake}, a large-scale multimodal deepfake benchmark for social-media content. It covers four modalities (image, audio, video, and audio-video talking heads) with over \textbf{1M} in-distribution samples from \textbf{30+} generation and manipulation methods in \textbf{Omni-Fake-Set}, and \textbf{200K+} out-of-distribution samples in \textbf{Omni-Fake-OOD} built from entirely disjoint generators, enabling realistic evaluation of generalization to unseen synthesis techniques. Both splits are annotated under a unified detection, localization, and explanation protocol with spatial and temporal labels. Building on this benchmark, we develop \textbf{Omni-Fake-R1}, a multimodal detector built upon Qwen2.5-Omni-7B~\citep{xu2025qwen2} and trained through a four-stage replay-based curriculum~\citep{wen2025light} combining SFT with Group Sequence Policy Optimization (GSPO)~\citep{zheng2025gspo}, which jointly optimizes all three tasks to produce consistent and interpretable outputs. The replay-based design preserves capabilities acquired in earlier stages, enabling effective cross-modal knowledge sharing as the model progressively learns new modalities.

In summary, our key contributions are threefold:

(1) We introduce \textbf{Omni-Fake}, the \textbf{first} unified four-modality deepfake benchmark for social media with over 1M in-distribution and 200K+ disjoint OOD samples, supporting joint evaluation of detection, localization, and explanation.

(2) We propose \textbf{Omni-Fake-R1}, a unified multimodal detection framework that combines curriculum SFT with modal replay and GSPO to jointly optimize detection, localization, and explanation, producing consistent and interpretable outputs across modalities.

(3) Extensive experiments show that Omni-Fake-R1 achieves state-of-the-art performance across all three tasks and generalizes well to unseen generators and platforms.

\section{Related Work}
\label{sec:related}

\noindent
\textbf{Deepfake Detection Benchmark.} Early datasets such as FaceForensics++~\citep{rossler2019faceforensics}, DFFD~\citep{dang2020detection}, and ForgeryNet~\citep{he2021forgerynet} laid the foundation for deepfake detection with large-scale paired real/fake samples, but mainly target facial forgeries and limited manipulation types~\citep{karras2021alias}. With diffusion and transformer-based generators, GenImage~\citep{zhu2023genimage} and DMImage~\citep{corvi2023detection} expanded to diverse AI-generated images. Socially grounded datasets such as SIDA~\citep{huang2025sida}, WildFake~\citep{hong2024wildfake}, Community Forensics~\citep{park2025community}, and So-Fake~\citep{huang2025so} capture real-world forgeries with richer annotations, yet still focus largely on visual content. Video benchmarks including DFDC~\citep{dolhansky2020deepfake}, AVCeleb~\citep{khalid2021fakeavceleb}, IDForge~\citep{xu2024identity}, and KoDF~\citep{kwon2021kodf} analyze temporal manipulations but seldom model audio–visual consistency. Recent multimodal benchmarks~\citep{zhou2021joint,zhao2025deepfakebench} combine visual and auditory cues, but remain fragmented and lack unified annotations for detection, localization, and explanation. LOKI~\citep{LOKI} evaluates multimodal models via QA-style detection and explanation, yet serves only as an evaluation suite. In contrast, Omni-Fake (Table~\ref{tab:dataset_comparison}) offers a substantially larger, training-ready four-modality corpus with pixel-level and temporal annotations, a unified detection–localization–explanation protocol, and a fully disjoint large-scale OOD split.
\begin{table*}[t]
\caption{Omni-Fake vs representative deepfake datasets. Compared with prior datasets, Omni-Fake provides a unified multimodal benchmark spanning image, audio, generic video, and audio-video talking-head inputs, supports multi-class detection with localization and explanation, and explicitly includes a held-out OOD split for generalization evaluation. (Multi-Mod. = multimodal, Multi-Cls. = multi-classification, Expl. = explanation)}
\centering
\footnotesize
\setlength{\tabcolsep}{4pt}
\begin{tabular}{lcccccccc}
\toprule
\textbf{Dataset} & \textbf{Modality} & \textbf{Social Media} &
\textbf{Multi-Mod.} & \textbf{MultiCls} & \textbf{Mask} &
\textbf{Expl.} & \textbf{OOD} & \textbf{Size} \\
\midrule
So-Fake~\citep{huang2025so}              & I              & \cmark & \xmark & \cmark & \cmark & \cmark & \cmark & 2M+ images \\
SID-Set~\citep{huang2025sida}              & I              & \cmark & \xmark & \cmark & \cmark & \cmark & \xmark & 0.3M images \\
TrueFake~\citep{dell2025truefake}             & I              & \cmark & \xmark & \xmark & \xmark & \xmark & \xmark & 0.6M social-media images \\
Fake2M~\citep{lu2023seeing}               & I              & \xmark & \xmark & \xmark & \xmark & \xmark & \xmark & 2M+ images \\
ForgeryNet~\citep{he2021forgerynet}           & I/V            & \xmark & \cmark & \cmark & \cmark & \xmark & \xmark & 2.9M images, 221K videos \\
DFDC~\citep{dolhansky2020deepfake}                 & V              & \xmark & \xmark & \xmark & \xmark & \xmark & \xmark & 128K videos \\
GenBuster-200K~\citep{wen2025busterx}       & V              & \xmark & \xmark & \xmark & \xmark & \xmark & \xmark & 200K short-form videos \\
AV-Deepfake1M~\citep{cai2024av}        & AV             & \xmark & \cmark & \xmark & \cmark & \xmark & \xmark & 1M+ AV clips \\
FakeAVCeleb~\citep{khalid2021fakeavceleb}          & AV             & \cmark & \cmark & \xmark & \xmark & \xmark & \xmark & 20K AV clips \\
Deepfake-Eval-2024~\citep{chandra2025deepfake}   & I/A/V          & \cmark & \cmark & \xmark & \xmark & \xmark & \xmark & 1{,}975 img; 45h V; 56.5h A \\
LOKI~\citep{LOKI}                 & I/A/V/T/3D     & \xmark & \cmark & \xmark & \xmark & \cmark & \xmark & $\approx$10K multimodal posts \\
\midrule
\textbf{Omni-Fake (Ours)} &
\textbf{I/A/V/AV-TH} &
\textbf{\cmark} & \textbf{\cmark} & \textbf{\cmark} &
\textbf{\cmark} & \textbf{\cmark} & \textbf{\cmark} &
\textbf{1M+ samples} \\
\bottomrule
\end{tabular}
\label{tab:dataset_comparison}
\end{table*}

\noindent
\textbf{Deepfake Detection Methods.}
Early deepfake detectors focused on image-level visual artifacts, using CNNs and handcrafted cues such as texture inconsistencies, blending boundaries, color shifts, and frequency-domain artifacts~\citep{rossler2019faceforensics,dang2020detection}. With the rise of vision-language models (VLMs), newer methods leverage CLIP-style embeddings, multi-view prompting, and semantic priors to distinguish real from synthetic content~\citep{huang2025sida,huang2025so,hong2024wildfake}.
Beyond static images, video-based approaches exploit temporal dynamics, modeling motion irregularities, temporal incoherence, and identity-manipulation trajectories~\citep{chen2024demamba,wen2025busterx,kundu2025towards,xu2024identity}, while audio-based detectors capture synthesized speech via spectral artifacts and speaker/prosody mismatches~\citep{li2024safeear,li2024cross}.
Recent multimodal methods integrate visual and auditory cues to detect lip-sync inconsistencies, semantic misalignment, and audio–visual desynchronization~\citep{liu2025beyond,anshul2025intra}. However, they usually provide limited detailed localization or explanation. They often produce sparse masks, weak evidence for specific regions, and shallow reasoning. They also lack a unified framework that jointly supports detection, localization, and explanation across modalities.

\noindent
\textbf{From Single-Modal Forensics to Unified Cross-Modal Learning.}
Early multimodal methods used shallow feature fusion with largely independent modality streams. Recent large multimodal models instead learn a shared semantic space linking visual, auditory, and textual signals~\citep{hurst2024gpt,alayrac2022flamingo,liu2024improved,li2023blip}, improving consistency and positive transfer across modalities. Models like Qwen2.5-Omni~\citep{xu2025qwen2} further show that jointly modeling vision, audio, and language yields stronger multimodal alignment. 
Our work follows this line by explicitly enforcing structured cross-modal alignment for coherent multimodal reasoning.

\noindent
\textbf{Post-training and Reinforcement Learning.} Training multimodal large language models (MLLMs) usually follows two stages: comprehensive multimodal pre-training, then post-training that combines supervised fine-tuning (SFT) with reinforcement learning with verifiable rewards (RLVR)~\citep{wen2025reinforcement}. Unlike reinforcement learning from human feedback (RLHF)~\citep{ouyang2022training} strategy, which optimizes for human preferences, RLVR uses objective, task-level signals as direct feedback. Algorithms such as PPO~\citep{schulman2017proximal}, DPO~\citep{rafailov2023direct}, and GRPO~\citep{shao2024deepseekmath} perform preference-based optimization, while GSPO extends RLVR to unified multimodal settings with structured rewards over textual, spatial, and temporal dimensions, promoting cross-modal consistency and reasoning alignment. 
We adopt GSPO within the RLVR paradigm to jointly optimize detection, localization, and explanation in a unified framework for multimodal deepfake detection.

\section{Dataset}
\label{sec:dataset}

\subsection{Overview}

Modern social media timelines mix authentic and AI-generated content across modalities, often after heavy re-encoding, editing, and cross-platform reposting. 
This makes multimodal deepfake detection far more challenging than conventional single-modality benchmarks with binary real/fake labels.

We introduce \textbf{Omni-Fake}, a unified multimodal dataset for social media deepfake detection, localization, and explanation.
It covers four modalities under a unified annotation protocol: images, audio, generic videos, and audio-visual talking-head videos. The benchmark consists of two complementary parts: \textbf{Omni-Fake-Set}, an in-distribution split (further divided into training and validation) for model development, and \textbf{Omni-Fake-OOD}, a benchmark split for evaluating generalization.
For images, audio, and generic videos, we adopt three labels: real, partially manipulated, and fully synthetic.
For talking-head videos, we use binary labels (real vs. fully synthetic) and focus on identity-driven and lip-driven face generation, as these are most closely tied to impersonation and fraud. 
Partially edited talking heads are evaluated under the generic video setting, where fine-grained spatial and temporal localization is supported.
Whenever available, we provide manipulation masks to enable unified evaluation across all three tasks.

\subsection{Data Collection}
\label{data collection}

\begin{figure}[t]  
  \centering
  \includegraphics[width=0.8\linewidth]{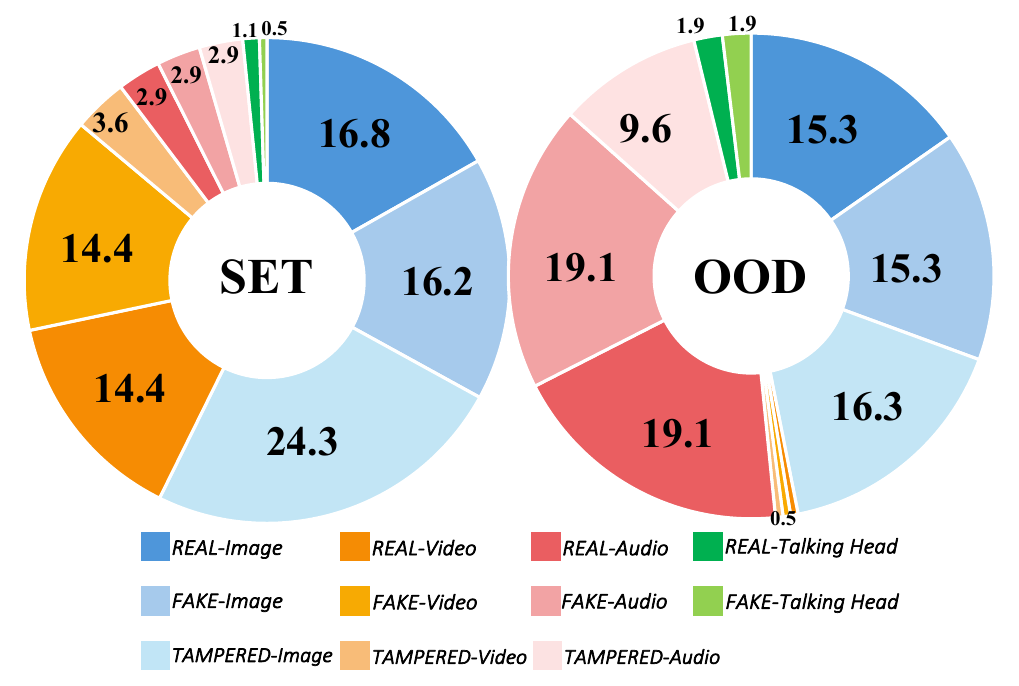}
  \caption{Composition of Omni-Fake across modalities and splits. The figure shows the label distribution of REAL, FULL SYNTHETIC, and TAMPERED samples in Omni-Fake-Set and Omni-Fake-OOD across image, audio, video, and AV talking-head data.}
  \label{fig:data_overview}
\end{figure}


\newcolumntype{L}[1]{>{\raggedright\arraybackslash}p{#1}}
\newcolumntype{C}[1]{>{\centering\arraybackslash}p{#1}}

\begin{figure*}[t]  
  \centering
  \includegraphics[width=0.92\linewidth]{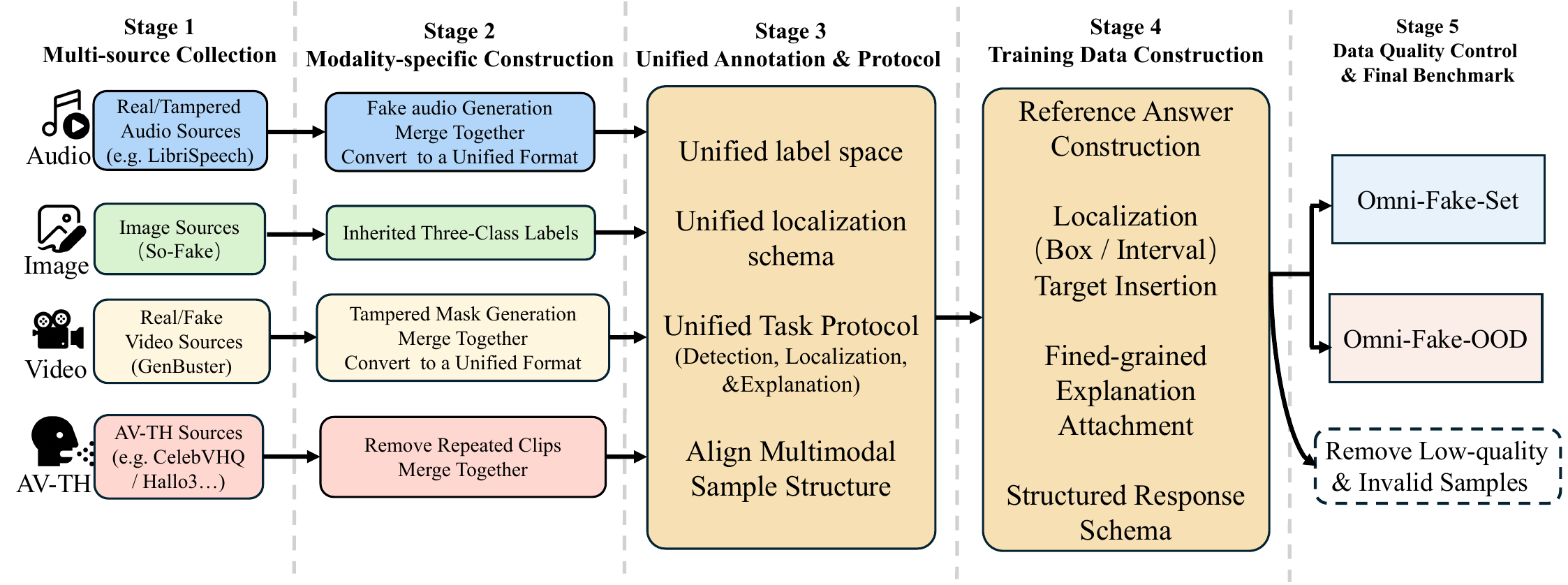}
 \caption{Data pipeline of Omni-Fake. Our pipeline not only unifies four modalities under a common protocol, but also builds high-quality training data and a dedicated OOD benchmark for trustworthy multimodal forgery evaluation.}
  \label{fig:semantic_alignment}
\end{figure*}

\begin{table}[t]
\caption{Data sources and representative generators in Omni-Fake.}
\centering
\scriptsize
\setlength{\tabcolsep}{2.5pt}
\renewcommand{\arraystretch}{1.05}

\begin{tabular}{L{0.82cm} C{0.56cm} C{3.15cm} L{2.95cm}}
\toprule
\textbf{Type} & \textbf{Split} & \textbf{Data sources} & \textbf{Representative generators} \\
\midrule

\multirow{2}{*}{Image}
& Set
& So-Fake-Set~\citep{huang2025so}.
& FLUX.1-dev~\citep{flux1dev2024}, Kandinsky3~\citep{arkhipkin2024kandinsky30technicalreport}, StyleGAN3\citep{karras2021alias}, and others. \\
\cmidrule(lr){2-4}
& OOD
& So-Fake-OOD~\citep{huang2025so}.
& GPT-4o~\citep{hurst2024gpt}, Ideogram3.0~\citep{ideogram2025ideogram3}, Nano Banana~\citep{google2025nanobanana}, and others. \\
\midrule

\multirow{2}{*}{Video}
& Set
& GenBuster-200K~\citep{wen2025busterx}, SENORITA-2M~\citep{zi2025se}, and others.
& CogVideoX~\cite{yang2025cogvideoxtexttovideodiffusionmodels}, EasyAnimate~\cite{xu2026easyanimatehighperformancevideogeneration}, HunyuanVideo~\citep{wen2025busterx}, and others. \\
\cmidrule(lr){2-4}
& OOD
& GenBuster-200K~\citep{wen2025busterx}, VideoPainter/VPBench~\citep{bian2025videopainter}, and others.
& Sora~\citep{sora-openai}, Pika~\citep{pika2024}, Gen3~\citep{runway_gen3_2024}, and others. \\
\midrule

\multirow{2}{*}{Audio}
& Set
& Multilingual LibriSpeech~\citep{pratap2020mls}, PartialEdit~\cite{Zhang_2025}, and others.
& Dia-1.6B~\citep{nari-labs-dia}, Kokoro-82M~\citep{hexgrad-kokoro82m}, Chatterbox~\citep{chatterboxtts2025}, and others. \\
\cmidrule(lr){2-4}
& OOD
& Common Voice~\cite{ardila2020commonvoicemassivelymultilingualspeech}, LlamaPartialSpoof~\citep{luong2025llamapartialspoof}, and others.
& Higgs-Audio~\citep{higgsaudio2025}, CosyVoice~\citep{du2024cosyvoice}, Fish Speech~\citep{liao2024fish}, and others. \\
\midrule

\multirow{2}{*}{AV-TH}
& Set
& celebVHQ~\citep{zhu2022celebv}, Hallo3~\citep{cui2025hallo3}, HDTF~\citep{zhang2021flow}, MAVOS~\citep{croitoru2025mavos}, TalkVid / TalkVid-bench~\citep{chen2025talkvid}, and others.
& AniPortrait~\citep{wei2024aniportrait}, EchoMimic~\citep{chen2024echomimic}, Hallo2~\citep{cui2024hallo2}, and others. \\
\cmidrule(lr){2-4}
& OOD
& FakeAVCeleb~\citep{khalid2021fakeavceleb}, TalkingHead-1KH~\citep{wang2021one}, and others.
& deepspeak-v2~\citep{barrington2024deepspeak}, Ditto~\citep{li2025ditto}, ACTalker~\citep{hong2025audio}, and others. \\
\bottomrule
\end{tabular}

\label{tab:data_sources_generators}
\end{table}
Omni-Fake integrates existing resources with newly collected multimodal forgeries under a unified label space (Figure~\ref{fig:semantic_alignment}). Omni-Fake-Set includes over 790K images, 210K videos, 120K audio clips, and 15K audio-visual talking-head videos; Omni-Fake-OOD includes 100K images, 3K videos, 100K audio clips, and 8K talking-head videos. The two splits are strictly disjoint in underlying content, speakers, data distributions, manipulation pipelines, and generative model families. No forgery method in Omni-Fake-OOD appears in Omni-Fake-Set.

\subsection{Overall Data Quality}
\label{sec:overall_quality}

A rigorous deepfake benchmark requires semantic coherence between its in-distribution and OOD splits, generator diversity, and high perceptual quality. We validate all three properties for Omni-Fake.

\noindent\textbf{Semantic Coherence.}Label proportions of real, partially manipulated, and fully synthetic samples are closely matched across Omni-Fake-Set and Omni-Fake-OOD in every modality, avoiding spurious gains from class imbalance. Figure~\ref{fig:data_overview} summarizes the composition of the benchmark across modalities and splits, showing that the two splits remain broadly aligned in label structure while differing in generator families and content sources.

\noindent\textbf{Generator Diversity.} Each modality combines multiple open-source and commercial synthesis and editing pipelines, with disjoint generator families assigned to Omni-Fake-Set and Omni-Fake-OOD wherever possible (Table~\ref{tab:data_sources_generators}). Per-family usage statistics (Sec.~\ref{data collection} and Appendix) show relatively even sampling, providing a realistic testbed for cross-generator generalization.

\noindent\textbf{Perceptual Quality.} We quantify realism using both automatic metrics and human evaluation. Table~\ref{tab:omni_fake_overall_quality} reports Fr'{e}chet distances, no-reference quality scores, mean opinion scores (MOS), and human real/fake detection accuracy. Omni-Fake-Set attains low distances and high MOS across modalities; Omni-Fake-OOD remains comparable in quality and is generally more challenging in most modalities.

\begin{table}[t]
    \caption{
        Overall data quality of \textbf{Omni-Fake-Set} and \textbf{Omni-Fake-OOD}.
        Lower values ($\downarrow$) indicate better distance / distortion or sync metrics;
        higher values ($\uparrow$) indicate better perceptual quality, intelligibility, or human detection performance. (MOS = Human Mean Opinion Score, HDA = Human Detection Accuracy)
    }
    \label{tab:omni_fake_overall_quality}
    \centering
    \scriptsize
    \setlength{\tabcolsep}{4pt}
    \begin{tabular}{llcccc}
        \toprule
        \textbf{Modality} & \textbf{Split} &
        \multicolumn{2}{c}{\textbf{Automatic quality / sync}} &
        \multicolumn{2}{c}{\textbf{Human evaluation}} \\
        \cmidrule(lr){3-4}\cmidrule(lr){5-6}

        \textbf{Image} & &
        \textbf{FID$\downarrow$} &
        \textbf{BRISQUE$\downarrow$} &
        \textbf{MOS$\uparrow$} &
        \textbf{HDA$\uparrow$} \\
        & Set &
        13.6 & 21.9 & 4.18 & 0.79 \\
        & OOD &
        19.2 & 26.7 & 3.93 & 0.72 \\

        \midrule

        \textbf{Video} & &
        \textbf{FVD$\downarrow$} &
        \textbf{BRISQUE$\downarrow$} &
        \textbf{MOS$\uparrow$} &
        \textbf{HDA$\uparrow$} \\
        & Set &
        137 & 24.8 & 4.11 & 0.83 \\
        & OOD &
        195 & 28.1 & 3.97 & 0.91 \\

        \midrule

        \textbf{Audio} & &
        \textbf{FAD$\downarrow$} &
        \textbf{PESQ$\uparrow$} &
        \textbf{MOS$\uparrow$} &
        \textbf{HDA$\uparrow$} \\
        & Set &
        1.6 & 4.20 & 4.27 & 0.62 \\
        & OOD &
        2.2 & 3.97 & 4.18 & 0.58 \\

        \midrule

        \textbf{Talking head} & &
        \textbf{FVD-AV$\downarrow$} &
        \textbf{LSE-C$\uparrow$ / LSE-D$\downarrow$} &
        \textbf{MOS$\uparrow$} &
        \textbf{HDA$\uparrow$} \\
        & Set &
        284 & 7.7\,/\,8.6 & 3.74 & 0.94 \\
        & OOD &
        272 & 7.3\,/\,9.0 & 3.27 & 0.89 \\

        \bottomrule
    \end{tabular}
\end{table}

\section{Method}
\label{sec:method}
\subsection{Overview}
\label{sec:method_overview}

We introduce \textbf{Omni-Fake-R1}, a unified multimodal baseline built on Qwen2.5-Omni-7B. Given any input (image, audio, video, or talking-head video), the model outputs a structured triple: a global authenticity label, spatial or temporal localization, and a textual rationale.

\begin{figure}[t]  
  \centering
  \includegraphics[width=1 \linewidth]{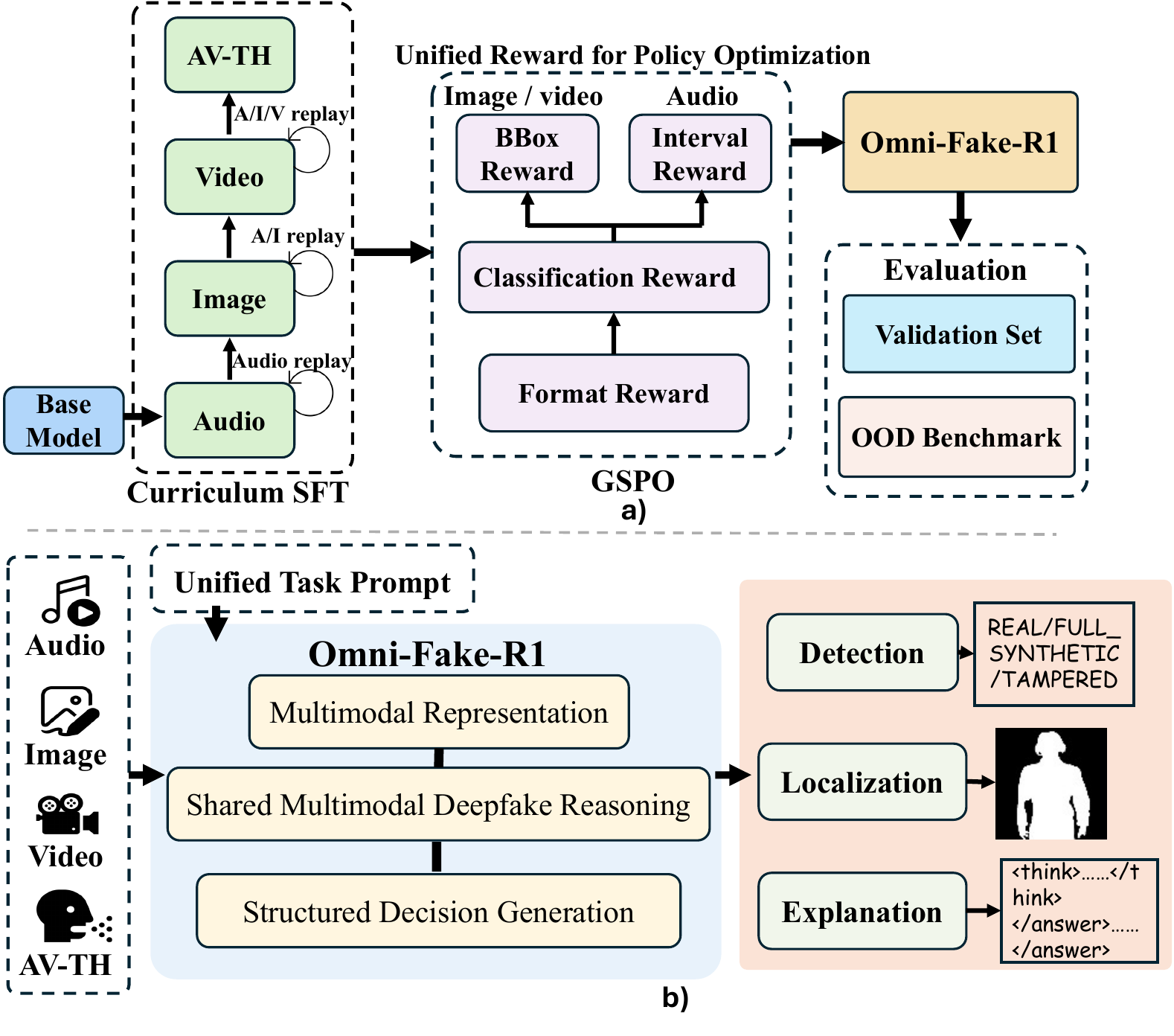}
  \caption{a) Training pipeline with SFT and GSPO. (b) Architecture of Omni-Fake-R1 producing detection, localization, and explanation outputs.} 
  \label{fig：main_fig.pdf}
\end{figure}

Training a single model to produce all three outputs across four modalities is challenging. Naively mixing modalities with standard log-likelihood training leads to task interference and poor metric optimization. We therefore adopt a two-stage pipeline (Figure~\ref{fig：main_fig.pdf}). In the first stage, curriculum SFT with modal replay introduces modalities incrementally, helping the model learn shared representations, follow the required output format, and resist catastrophic forgetting. In the second stage, unified GSPO reinforcement learning optimizes task-level rewards for classification, localization, and explanation jointly, yielding more balanced and robust predictions under OOD settings.

\subsection{Training}
\label{subsec:training}

\subsubsection{Curriculum SFT with Modal Replay}


Introducing all four modalities simultaneously causes earlier skills to be overwritten by modalities with larger data volumes. We therefore train in a four-stage curriculum that adds one modality at a time: audio, then images, then videos, then talking-head videos.

At each stage, the full training set of the new modality is mixed with a 15\% replay subset from every previously seen modality. This simple schedule prevents catastrophic forgetting at low overhead and allows later modalities to benefit from representations learned in earlier stages.

\subsubsection{Unified GSPO Reinforcement Learning}

Curriculum SFT yields a strong starting point, but it still optimizes
next-token likelihood rather than our task metrics: authenticity,
localization and explanation quality. To directly align the model with
our detection--localization--explanation protocol, we add a unified GSPO
reinforcement learning phase on top of the SFT checkpoint. Following
GSPO ~\citep{zheng2025gspo}, we sample multiple responses per
input from any modality, score each response with a scalar
detection--localization--explanation reward $r(x,y)$ (Section~\ref{subsec:rl_rewards}),
and update the policy using group-wise relative advantages with a KL
penalty to stay close to the SFT model. Concretely, GSPO
maximizes
\small
\begin{equation}
\begin{aligned}
  J_{\text{GSPO}}(\theta)
  &= \mathbb{E}_{x,\{y_i\}}
  \Bigg[
    \frac{1}{G} \sum_{i=1}^G
    \frac{1}{|y_i|}
    \sum_{t=1}^{|y_i|}
    \min\big(
      s_{i,t}(\theta)\,\hat{A}_{i,t}, \\
  &\qquad\qquad\qquad
      \mathrm{clip}(s_{i,t}(\theta),1-\epsilon,1+\epsilon)\,\hat{A}_{i,t}
    \big)
  \Bigg],
\end{aligned}
\end{equation}
where $\hat{A}_{i,t}$ is a token-level advantage and $s_{i,t}(\theta)$
is the corresponding importance ratio; we follow
\citet{zheng2025gspo} for the exact definitions.

Intuitively, this encourages responses with correct labels, masks or intervals, and output format, while enabling stable token-level updates under the detect--locate--explain reward. Consequently, the second stage jointly sharpens detection, localization, and explanation, and improves robustness across modalities and out-of-distribution generators.

\subsection{Reinforcement learning rewards design}
\label{subsec:rl_rewards}

To align the model with our unified detection-localization-explanation protocol, we use a single scalar reward that combines four terms:
format, detection, spatial localization, temporal localization:
\[
  r(x,y)
  =
  \lambda_{\mathrm{fmt}} r_{\mathrm{fmt}}
  + \lambda_{\mathrm{acc}} r_{\mathrm{acc}}
  + \lambda_{\mathrm{bbox}} r_{\mathrm{bbox}}
  + \lambda_{\mathrm{int}} r_{\mathrm{int}},
\]
where the weights will be reported in the appendix.

\noindent
\textbf{Format Reward.} A deterministic parser verifies that there is exactly one pair of
\texttt{<think>}~\dots~\texttt{</think>} tags and one pair of
\texttt{<answer>}~\dots~\texttt{</answer>} tags, and that it can extract
a valid label and, when present, well-formed
\texttt{<|box\_start|>}~\dots~\texttt{<|box\_end|>} or
\texttt{<|interval\_start|>}~\dots~\texttt{<|interval\_end|>}
segments from the \texttt{<answer>} block. Responses that pass all
these checks receive a format reward of 1, otherwise 0. 
This encourages parseable reasoning traces and enables reliable extraction of labels, masks, and intervals for other rewards. 

\noindent
\textbf{Detection Reward.}
The detection term compares the predicted authenticity label in the
\texttt{<answer>} block with the ground truth. Since tampered cases are
usually more subtle than real or fully synthetic samples, we assign a
larger positive reward to correct \texttt{TAMPERED} predictions and a
smaller reward to correct \texttt{REAL} and \texttt{FULL\_SYNTHETIC}
predictions, and give zero reward to incorrect labels. This prevents
optimization from being dominated by easier classes, and encourages
balanced performance across the ternary label space.

\noindent
\textbf{Spatial Localization Reward.}
For images and videos with pixel-level manipulation masks, we derive a
ground-truth bounding box from the mask and compare it with the
predicted box in the \texttt{<answer>} block. For tampered samples, the
spatial reward is the Intersection over Union (IoU) between the
predicted box and the ground-truth box. For real and fully synthetic
samples, the reward is 1 if the model predicts no boxes and 0
otherwise. In this way, tampered cases are rewarded for accurate box
localization, while genuine samples are rewarded for correctly
predicting the absence of manipulated regions.

\noindent
\textbf{Temporal Localization Reward.} For audio and video samples with annotated forged intervals, we parse
the predicted intervals from the \texttt{<answer>} block and compare
them to the ground truth intervals using a one-dimensional IoU measure
on the time axis. We perform bipartite matching between predicted and
true intervals and average the IoU over matched pairs. As in the spatial
case, tampered samples are rewarded for precise interval prediction,
while real and fully synthetic samples are rewarded only when the model
does not output spurious intervals.

\section{Experiments}
\label{sec:exp}

We evaluate Omni-Fake-R1 on Omni-Fake-Set,
Omni-Fake-OOD.
The experiments are designed to:
(i) Compare a single unified model against strong single-modality
baselines on audio, image, video, and AV talking-head tasks;
(ii) Assess the behaviour of representative models under
out-of-distribution (OOD) evaluation;
(iii) Measure the robustness of Omni-Fake-R1 to common post-processing
operations and corruptions;
(iv) Quantify the impact of different supervised fine-tuning strategies
and unified GSPO reinforcement learning on overall performance;
(v) Evaluate the quality of model explanations in comparison with
baselines.

\subsection{Experimental setup}
\label{subsec:setup}

\paragraph{Baselines.}
For each modality, we compare Omni-Fake-R1 with representative detection-only models, localization-oriented models, and
vision–language models. On images, we include CnnSpot~\citep{wang2020cnn}, UnivFD~\citep{ojha2023towards},
AntiFakePrompt~\citep{chang2023antifakeprompt}, HIFI-Net~\citep{guo2025language}, SIDA~\citep{huang2025sida}, LLaVA-1.5-13B~\citep{liu2024improved} and DeepSeek-VL-7B~\citep{lu2024deepseek}.
On videos, we evaluate 3D ResNeXt, VideoMAE~\citep{tong2022videomae}, DeMamba~\citep{chen2024demamba}, VideoLISA~\citep{bai2024one},
Qwen2.5-VL-7B~\citep{xu2025qwen2} and InternVL3-8B~\citep{zhu2025internvl3exploringadvancedtraining}. On audio, we compare to AASIST~\citep{jung2022aasist},
SSL-AASIST~\citep{tak2022automatic}, SafeEar~\citep{li2024safeear}, PartialSpoof~\citep{zhang2022partialspoof} and FakeSound~\citep{xie2024fakesound}. On AV talking-head
clips, we include LipForensics~\citep{haliassos2021lips}, LIPINC~\citep{datta2024exposing}, RealForensics~\citep{haliassos2022leveraging} and AVH-Align~\citep{smeu2025circumventing}.
All baselines are fine-tuned on
Omni-Fake-Set with their recommended hyperparameters.

\paragraph{Metrics.}
We use accuracy (Acc) and macro F1 for detection, and mask IoU and localization F1 for localization.
For audio, we compute interval IoU and F1. Explanation quality is evaluated using ROUGE-L and Cosine Semantic Similarity (CSS) between model explanations and reference rationales. 
For human evaluation, we collect ratings from ten domain experts on a 5-point Likert scale, scoring factual correctness and usefulness for non-expert auditors.

\subsection{Single-modality results}
\label{subsec:single_modal}

Table~\ref{tab:Set_result} reports single-modality
results on the image, video, audio and AV talking-head subsets of
Omni-Fake. On images, Omni-Fake-R1
achieves the best overall balance, matching or surpassing IFDL
baselines in detection while further improving spatial localization.
On generic videos, Omni-Fake-R1
outperforms clip-level detectors and VLM style video models in both
detection F1 and temporal localization. On audio, Omni-Fake-R1 attains the highest ternary
detection accuracy and the best interval-level localization, clearly
outperforming recent spoofing and partial-manipulation baselines.
On AV talking-head videos, Omni-Fake-R1
achieves the best fake-detection F1 over specialized AV deepfake and
lip-sync detectors, while maintaining the same structured tag-based
output as in the single modality settings.

\begin{table}[t]
\caption{Performance comparisons on
Validation set of Omni-Fake-Set across four modalities. (AV-TH = audio--video
talking-head videos).}
\label{tab:Set_result}
\centering
\tiny
\setlength{\tabcolsep}{1pt}
\renewcommand{\arraystretch}{0.8}
\resizebox{0.95\columnwidth}{!}{%
\begin{tabular}{llcccc}
\toprule
\textbf{Mod.} & \textbf{Method} &
\multicolumn{2}{c}{\textbf{Detection}} &
\multicolumn{2}{c}{\textbf{Localization}} \\
\cmidrule(lr){3-4} \cmidrule(lr){5-6}
& & Acc & F1 & IoU & F1 \\
\midrule
\multirow{8}{*}{Image}
  & CnnSpot~\citep{wang2020cnn}                  & 87.12 & 85.38 & --    & --    \\
  & UnivFD~\citep{ojha2023towards}              & 81.57 & 61.91 & --    & --    \\
  & AntiFakePrompt~\citep{chang2023antifakeprompt} & 88.04 & 86.86 & --    & --    \\
  & FakeVLM~\citep{wen2025spot}                  & 90.34 & 89.21 & -- & -- \\
  & LEGION~\citep{kang2025legion}                  & 88.38 & 87.41 & -- & -- \\
  & HIFI-Net~\citep{guo2025language}            & 37.23 & 23.64 & 13.79 & 17.31 \\
  & SIDA~\citep{huang2025sida}                  & 89.88 & 88.15 & 46.27 & 56.10 \\
  & LLaVA-1.5-13B~\citep{liu2024improved}  & 81.69 & 80.11 & 27.36 & 39.59 \\
  & DeepSeek-VL-7B~\citep{lu2024deepseek}       & 80.93 & 79.08 & 33.82 & 41.43 \\
  & \textbf{Ours}                               & \textbf{91.92} & \textbf{90.58} &
                                                   \textbf{47.06} & \textbf{51.63} \\
\midrule
\multirow{7}{*}{Video}
  & 3D ResNeXt~\citep{hara2018can}                                  & 72.48 & 74.51 & --    & --    \\
  & VideoMAE~\citep{tong2022videomae}          & 79.34 & 81.09 & --    & --    \\
  & DeMamba~\citep{chen2024demamba}            & 84.22 & 82.19 & --    & --    \\
  & VideoLISA~\citep{bai2024one}               & 66.41 & 64.58 & 23.14 & 27.63 \\
  & Qwen2.5-VL-7B~\citep{xu2025qwen2}          & 76.45 & 74.19 & 33.14 & 35.98 \\
  & InternVL3-8B~\citep{zhu2025internvl3exploringadvancedtraining} & 71.63 & 73.26 & 32.37 & 39.01 \\
  & \textbf{Ours}                              & \textbf{89.84} & \textbf{88.29} &
                                                  \textbf{40.63} & \textbf{43.35} \\
\midrule
\multirow{6}{*}{Audio}
  & AASIST~\citep{jung2022aasist}              & 68.73 & 65.31 & --    & --    \\
  & SSL-AASIST~\citep{tak2022automatic}        & 72.94 & 70.86 & --    & --    \\
  & SafeEar~\citep{li2024safeear}              & 81.62 & 79.27 & --    & --    \\
  & PartialSpoof~\citep{zhang2022partialspoof} & 63.82 & 61.74 & 18.26 & 21.03 \\
  & FakeSound~\citep{xie2024fakesound}        & 75.41 & 73.92 & 26.71 & 29.58 \\
  & \textbf{Ours}                              & \textbf{92.13} & \textbf{90.47} &
                                                  \textbf{45.92} & \textbf{47.58} \\
\midrule
\multirow{6}{*}{AV-TH}
  & LipForensics~\citep{haliassos2021lips}     & 82.73 & 81.46 & --    & --    \\
  & AVAD~\citep{feng2023self}                                        & 74.28 & 72.14 & --    & --    \\
  & LIPINC~\citep{datta2024exposing}          & 86.94 & 85.21 & --    & --    \\
  & RealForensics~\citep{haliassos2022leveraging} & 88.17 & 86.73 & --    & --    \\
  & AVH-Align~\citep{smeu2025circumventing}   & 78.42 & 76.28 & --    & --    \\
  & \textbf{Ours}                              & \textbf{96.18} & \textbf{95.54} &
                                                  -- & -- \\
\bottomrule
\end{tabular}%
}
\end{table}

\subsection{Out-of-distribution generalization}
\label{subsec:ood}

We examine OOD performance using Omni-Fake-OOD. For each modality, we select three representative state-of-the-art models: a strong detection-only baseline, a VLM baseline, and Omni-Fake-R1. Table~\ref{tab:ood_result} summarizes the results.

Across all modalities, performance drops when moving from
Omni-Fake-Set to Omni-Fake-OOD, reflecting shifts in generators,
content sources, and post-processing. Detection-only baselines degrade the most, especially on partial manipulations. VLM models are more stable but still lose a large fraction of localization performance. Omni-Fake-R1 consistently achieves the highest OOD Acc/F1 and IoU/F1, with particularly strong gains on AV tasks. This suggests that the unified detection–localization–explanation protocol and multimodal curriculum help the model rely on semantic and cross-modal inconsistencies rather than generator-specific artefacts.

\begin{table}[t]
\caption{Performance comparisons on
Omni-Fake-OOD across four modalities. (AV-TH = audio--video
talking-head videos.)}
\label{tab:ood_result}
\centering
\tiny
\setlength{\tabcolsep}{1pt}
\renewcommand{\arraystretch}{0.8}
\resizebox{0.95\columnwidth}{!}{%
\begin{tabular}{llcccc}
\toprule
\textbf{Mod.} & \textbf{Method} &
\multicolumn{2}{c}{\textbf{Detection}} &
\multicolumn{2}{c}{\textbf{Localization}} \\
\cmidrule(lr){3-4} \cmidrule(lr){5-6}
& & Acc & F1 & IoU & F1 \\
\midrule
\multirow{4}{*}{Image}
  & CnnSpot~\citep{wang2020cnn}        & 67.78 & 65.91 & --    & --    \\
  & SIDA~\citep{huang2025sida}           & 74.24 & 77.60 & 36.32 & 41.67 \\
  & DeepSeek-VL-7B~\citep{lu2024deepseek} & 66.41 & 63.19 & 23.07 & 30.15 \\
  & \textbf{Ours}  & \textbf{79.25} & \textbf{77.71} &
                     \textbf{37.54} & \textbf{43.63} \\
\midrule
\multirow{4}{*}{Video}
  & 3D ResNeXt~\citep{hara2018can}     & 64.03 & 61.29 & --    & --    \\
  & DeMamba~\citep{chen2024demamba}        & 78.93 & 74.74 & --    & --    \\
  & Qwen2.5-VL-7B~\citep{xu2025qwen2}  & 70.43 & 68.98 & 24.89 & 22.57 \\
  & \textbf{Ours}  & \textbf{85.96} & \textbf{82.53} &
                     \textbf{34.20} & \textbf{39.51} \\
\midrule
\multirow{4}{*}{Audio}
  & AASIST~\citep{jung2022aasist}         & 57.14 & 55.09 & --    & --    \\
  & SafeEar~\citep{li2024safeear}        & 71.29 & 68.96 & --    & --    \\
  & FakeSound~\citep{xie2024fakesound}      & 64.51 & 65.71 & 10.33 & 16.02 \\
  & \textbf{Ours}  & \textbf{83.85} & \textbf{82.10} &
                     \textbf{31.93} & \textbf{33.86} \\
\midrule
\multirow{4}{*}{AV-TH}
  & LipForensics~\citep{haliassos2021lips}   & 73.52 & 70.13 & --    & --    \\
  & RealForensics~\citep{haliassos2022leveraging}  & 72.96 & 71.62 & --    & --    \\
  & AVH-Align~\citep{smeu2025circumventing}      & 59.06 & 64.06 & --    & --    \\
  & \textbf{Ours}  & \textbf{86.52} & \textbf{86.20} &
                     -- & -- \\
\bottomrule
\end{tabular}%
}
\end{table}

\subsection{Robustness evaluation}
\label{subsec:robustness}

To mimic real social-media pipelines, we apply common channel
corruptions to Omni-Fake-OOD, including JPEG compression, blur,
additive noise, random cropping and rescaling, and codec
re-encoding. We then evaluate Omni-Fake-R1 under each corruption and
under the original clean setting (``Ours'') using the same unified
metrics.
Table~\ref{tab:robustness} reports modality-averaged scores, obtained
by averaging detection metrics over all four modalities and localization metrics over the three
modalities with spatial/temporal annotations (image, video and audio).

As expected, performance degrades under stronger corruptions, but
Omni-Fake-R1 maintains high detection F1 and localization IoU across
all settings, suggesting that the unified SFT + GSPO training
confers robustness to realistic channel effects.

\begin{table}[t]
\caption{Robustness of the unified SFT + GSPO model under
common corruptions. Metrics are averaged over all modalities.}
\centering
\setlength{\tabcolsep}{3pt}
\begin{tabular}{lcccc}
\toprule
\textbf{Setting} &
\multicolumn{2}{c}{\textbf{Detection}} &
\multicolumn{2}{c}{\textbf{Localization}} \\
\cmidrule(lr){2-3} \cmidrule(lr){4-5}
& Acc & F1 & IoU & F1 \\
\midrule
JPEG 70     & 89.92 & 88.94 & 40.05 & 41.29 \\
JPEG 80     & 91.57 & 89.83 & 41.41 & 43.76 \\
Resize 0.5  & 90.28 & 88.65 & 40.94 & 40.32 \\
Gaussian 10 & 88.46 & 87.72 & 39.67 & 41.85 \\
Ours    & \textbf{92.52} & \textbf{91.22} &
              \textbf{44.54} & \textbf{47.52} \\
\bottomrule
\end{tabular}
\label{tab:robustness}
\end{table}

\subsection{Ablation studies}
\label{subsec:ablation}

\noindent
\textbf{Supervised fine-tuning strategies.} We compare three SFT strategies using the same data and backbone:
(i) \emph{Single-modality SFT}, which trains four independent models
for audio, image, video and AV;
(ii) \emph{Full-mix SFT}, which trains a single model on mixed
A+I+V+AV batches from the beginning; and
(iii) \emph{Curriculum SFT}, which gradually unlocks modalities
following the sequence A $\rightarrow$ AI $\rightarrow$ AIV
$\rightarrow$ AIV-AV.
As shown in Table~\ref{tab:training_ablation}, single-modality SFT gives strong per-modality detection but no unified model, and full-mix SFT suffers from modality imbalance. Curriculum SFT matches or slightly improves single-modality detection while providing best localization and OOD performance.

\noindent
\textbf{Replay ratio.} Within curriculum SFT, we sweep replay ratios for earlier-stage data (0\%, 5\%, 10\%, 15\%, 30\%). Very small ratios (below 5\%) slow forgetting but cannot prevent it, while a large ratio (30\%) preserves early modalities yet hinders learning new ones and can cause negative transfer. Ratios around 10--15\% offer the best trade-off, so we adopt 15\% in all experiments. (See Appendix for more details)

\noindent
\textbf{Unified GSPO reinforcement learning.} We compare three variants: an \emph{SFT-only} model without RL, an \emph{RL-only} model that applies GSPO updates without curriculum SFT, and our default \emph{unified GSPO RL} model that combines curriculum SFT with the full multi-term reward. As shown in Table~\ref{tab:training_ablation}, the RL-only variant performs poorly on both detection and localization and is clearly inferior to any SFT-based model. Unified GSPO RL, on top of SFT, consistently improves detection and spatial/temporal localization across modalities, confirming the benefit of the full detect–locate training scheme.

\begin{table}[t]
\caption{
Ablation on training strategies, including supervised fine-tuning and unified GSPO-token RL.
}
\centering
\footnotesize
\setlength{\tabcolsep}{3pt}
\begin{tabular}{llcccc}
\toprule
\textbf{Method} & \textbf{Modality} &
\multicolumn{2}{c}{\textbf{Det.}} &
\multicolumn{2}{c}{\textbf{Loc.}} \\
\cmidrule(lr){3-4} \cmidrule(lr){5-6}
& & Acc & F1 & IoU & F1 \\
\midrule

\multirow{4}{*}{Single-mod SFT}
  & Audio        & 79.73 & 78.42 & 37.36 & 37.08 \\
  & Image        & 83.76 & 84.95 & 41.15 & 45.83 \\
  & Video        & 76.81 & 71.33 & 36.49 & 35.21 \\
  & AV-TH        & 89.74 & 90.57 & -- & -- \\
\midrule

\multirow{4}{*}{Full-mix SFT}
  & Audio        & 71.96 & 69.81 & 25.53 & 24.71 \\
  & Image        & 79.29 & 78.75 & 36.86 & 39.42 \\
  & Video        & 69.57 & 68.40 & 24.69 & 26.37 \\
  & AV-TH        & 82.63 & 83.07 & -- & -- \\
\midrule

\multirow{4}{*}{Curriculum SFT}
  & Audio        & 88.45 & 86.27 & 42.01 & 43.58 \\
  & Image        & 87.79 & 87.86 & 45.61 & 43.94 \\
  & Video        & 83.43 & 85.70 & 38.25 & 37.49 \\
  & AV-TH        & 91.33 & 90.07 & -- & -- \\
\midrule

\multirow{4}{*}{\shortstack[c]{\textbf{SFT + Unified}\\\textbf{GSPO-token RL}}}
  & \textbf{Audio}   & \textbf{92.13} & \textbf{90.47} & \textbf{45.92} & \textbf{47.58} \\
  & \textbf{Image}   & \textbf{91.92} & \textbf{90.58} & \textbf{47.06} & \textbf{51.63} \\
  & \textbf{Video}   & \textbf{89.84} & \textbf{88.29} & \textbf{40.63} & \textbf{43.35} \\
  & \textbf{AV-TH}   & \textbf{96.18} & \textbf{95.54} & \textbf{--} & \textbf{--} \\
\bottomrule
\end{tabular}
\label{tab:training_ablation}
\end{table}

\subsection{Explanation study}
\label{subsec:explanation}

We assess explanation quality with \textbf{automatic metrics and human
judgment}. Full results are given in the supplementary material.
Across all four modalities, Omni-Fake-R1 achieves the best ROUGE-L and
CSS scores. Removing the explanation-related terms from the
GSPO reward substantially reduces CSS while leaving detection
almost unchanged, showing that RL mainly shapes the rationales rather
than the labels.
For human evaluation, ten experts rate sampled instances per
modality on factual correctness and usefulness for non-expert auditors.
Omni-Fake-R1 obtains the highest mean scores and the lowest variance,
and its explanations more frequently align with ground-truth regions or
intervals. Human scores correlate well with CSS, supporting CSS as a
practical automatic proxy for explanation quality.

\section{Conclusion}
\label{sec:conclusion}

We propose Omni-Fake, a new benchmark for multimodal deepfake detection covering images, audio, video, and talking heads. The benchmark includes large-scale in-distribution data and an OOD suite, enabling rigorous evaluation of robustness and cross-modal generalization. We further present an RL-driven multimodal detector that improves cross-modal reasoning and delivers strong gains in detection, localization, and explanation. Together, Omni-Fake and our detector provide a solid foundation for advancing real-world multimodal misinformation forensics.

\noindent
\textbf{Acknowledgements}
This work is supported by The Alan Turing Institute (UK) through the project ''Turing-DSO Labs Singapore Collaboration'' (SDCfP2 \textbackslash 100009) and EPSRC IAA Grant (175944).

\clearpage

{
    \small
    \bibliographystyle{ieeenat_fullname}
    \bibliography{reference}
}
\clearpage
\setcounter{page}{1}
\maketitlesupplementary

\section*{Overview of Supplementary Material}
\label{sec:supp-overview}

In this supplementary document, we provide:
\begin{itemize}
    \item \textbf{S1.~Extended Experimental Results:} extra quantitative results, including explanation quality, replay ratio ablation, and RL reward design.
    \item \textbf{S2.~Dataset statistics:} core statistics of the \textsc{Omni-Fake} benchmark across modalities and splits.
    \item \textbf{S3.~Implementation Settings:} key details of the training setup for SFT and GSPO-based RL.
    \item \textbf{S4.~Case Studies and Representative Samples:} qualitative examples with visualizations of masks, intervals, textual explanations, and representative samples from \textsc{Omni-Fake-Set} and \textsc{Omni-Fake-OOD} across all modalities.
\end{itemize}


\section*{S1.~Extended Experimental Results}
\label{sec:s1-extended}

We report additional experiments on explanation quality, the effect of replay ratios during multimodal training, and the RL reward used for alignment.


\subsection*{S1.1~Explanation Study}
\label{sec:s1.1-explanation}

We evaluate explanations using ROUGE-L (longest common subsequence F-measure), cosine semantic similarity (CSS) between sentence embeddings, and human expert ratings on a 1--5 scale for factual correctness and usefulness. ROUGE-L reflects lexical and structural overlap, CSS captures semantic similarity, and human scores provide a direct assessment of explanation quality.

Table~\ref{tab:exp-explanation-metrics} reports results for images, audio, generic videos, and audio--visual talking-head clips. CSS is high across modalities, ROUGE-L is moderate due to paraphrasing, and human scores are above 4 on average, indicating that explanations are generally accurate and informative.

\begin{table}[H]
  \centering
  \caption{Explanation quality across modalities.}
  \label{tab:exp-explanation-metrics}
  \begin{tabular}{lccc}
    \toprule
    & \multicolumn{2}{c}{Automatic metrics} & Human experts \\
    \cmidrule(lr){2-3}\cmidrule(lr){4-4}
    Modality 
    & ROUGE-L ($\uparrow$) 
    & CSS ($\uparrow$) 
    & Mean score ($\uparrow$) \\
    \midrule
    Image   & 0.41 & 0.79 & 4.39 \\
    Audio   & 0.32 & 0.70 & 4.27 \\
    Video   & 0.37 & 0.72 & 4.21 \\
    AV-TH   & 0.45 & 0.86 & 4.52 \\
    \bottomrule
  \end{tabular}
\end{table}


\subsection*{S1.2~Replay Ratio Ablation}
\label{sec:s1.2-replay}

We study replay in a two-modality curriculum (Audio $\rightarrow$ Image). The model is first trained on audio only, then on images while replaying a proportion $p \in \{0\%, 5\%, 10\%, 15\%, 30\%\}$ of audio data. We evaluate the final model with the average detection ACC and average localization IoU over both modalities.

Table~\ref{tab:replay-ablation} shows that very small replay (0--5\%) slows but does not prevent forgetting; a large ratio (30\%) protects early modalities but harms learning of later ones; ratios around 10--15\% give the best trade-off. We therefore use 15\% replay in all experiments.

\begin{table}[t]
  \centering
  \caption{Replay ratio ablation on the Audio→Image two-modality training setup.} 
  
  \label{tab:replay-ablation}
  \begin{tabular}{lcc}
    \toprule
    \textbf{Replay Ratio} & \textbf{Avg.\ ACC ($\uparrow$)} & \textbf{Avg.\ IoU ($\uparrow$)} \\
    \midrule
    0\%   & 70.19 & 31.72 \\
    5\%   & 75.35 & 33.42 \\
    10\%  & 78.14 & 36.07 \\
    \textbf{15\%}  & \textbf{80.45} & \textbf{39.49} \\
    30\%  & 79.86 & 37.31 \\
    \bottomrule
  \end{tabular}
\end{table}


\subsection*{S1.3~RL Reward Design}
\label{sec:s1.3-rl-reward}

In the RL alignment stage, we optimize a composite reward
\begin{equation}
    r(x, y) = 
    \lambda_{\text{fmt}} r_{\text{fmt}} +
    \lambda_{\text{acc}} r_{\text{acc}} +
    \lambda_{\text{bbox}} r_{\text{bbox}} +
    \lambda_{\text{int}} r_{\text{int}},
\end{equation}
where:
\begin{itemize}
    \item $r_{\text{fmt}}$ checks output format \verb|<think>| and \verb|<answer>| tags and field validity.
    \item $r_{\text{acc}}$ measures global classification correctness for REAL / TAMPERED / FULLY SYNTHETIC or REAL / FAKE.
    \item $r_{\text{bbox}}$ scores spatial localization via box IoU.
    \item $r_{\text{int}}$ scores temporal localization via interval IoU.
\end{itemize}

We set
\[
    \lambda_{\text{fmt}} = 0.3,\quad
    \lambda_{\text{acc}} = 0.5,\quad
    \lambda_{\text{bbox}} = 1.0,\quad
    \lambda_{\text{int}} = 1.0,
\]
balancing structural correctness and global decisions, while putting stronger weight on localization quality. This configuration yields stable RL training and consistent gains in detection and localization.

\begin{table*}[t]
  \centering

  \caption{Table S1.1: Number of samples per modality, label type, and split in \textsc{Omni-Fake}. 
  The left block shows counts for the in-distribution \textsc{Omni-Fake-Set}, while the right block
  shows counts for the out-of-distribution \textsc{Omni-Fake-OOD}.}
  \label{tab:s1.1-overall}

  \resizebox{\textwidth}{!}{
  \begin{tabular}{lrrrrrrrr}
    \toprule
    & \multicolumn{4}{c}{\textsc{Omni-Fake-Set}} 
    & \multicolumn{4}{c}{\textsc{Omni-Fake-OOD}} \\
    \cmidrule(lr){2-5} \cmidrule(lr){6-9}
    Modality 
    & REAL & FULLY SYNTHETIC & TAMPERED & Total 
    & REAL & FULLY SYNTHETIC & TAMPERED & Total \\
    \midrule
    Image 
    & 232{,}000 & 224{,}000 & 336{,}000 & 792{,}000
    & 32{,}000 & 32{,}000 & 34{,}000 & 98{,}000 \\
    
    Audio 
    & 40{,}000 & 40{,}000 & 40{,}000 & 120{,}000
    & 40{,}000 & 40{,}000 & 20{,}000 & 100{,}000 \\
    
    Video 
    & 100{,}000 & 100{,}000 & 10{,}000 & 210{,}000
    & 1{,}000 & 1{,}000 & 1{,}000 & 3{,}000 \\
    
    AV-TH
    & 7{,}500 & 7{,}500 & 0 & 15{,}000
    & 4{,}000 & 4{,}000 & 0 & 8{,}000 \\
    
    \midrule
    Total 
    & 379{,}500 & 371{,}500 & 386{,}000 & 1{,}137{,}000
    & 77{,}000 & 77{,}000 & 55{,}000 & 209{,}000 \\
    \bottomrule
  \end{tabular}
  }

\end{table*}

\section*{S2.~Dataset Statistics}
\label{sec:s2-dataset}

We summarize core statistics of \textsc{Omni-Fake} across four modalities (images, audio, videos, and audio--visual talking-head clips) and three label types (REAL, FULLY SYNTHETIC, TAMPERED), for both the in-distribution \textsc{Omni-Fake-Set} and out-of-distribution \textsc{Omni-Fake-OOD} splits.

These statistics show that \textsc{Omni-Fake} is large-scale, spans multiple modalities and manipulation types, and includes a substantial OOD split, making it suitable for evaluating unified multimodal deepfake detectors under distribution shifts.


\section*{S3.~Implementation Settings}
\label{sec:s3-impl}

All experiments are conducted on a single node with 4$\times$NVIDIA H20 96GB GPUs using
PyTorch, DeepSpeed ZeRO-2 and FlashAttention-2. Our base model is Qwen/Qwen2.5-Omni-7B, which
is first fine-tuned with LoRA rank 16, $\alpha=32$, dropout 0.05 on the merged Omni-Fake SFT
dataset. We then apply GSPO-based reinforcement learning on the RL-formatted multimodal data,
using the composite reward described in Section~\ref{subsec:rl_rewards}. Hyperparameters
follow standard large-model training practice and emphasize stability rather than aggressive
tuning. The complete training process requires approximately 100 GPU-hours on the H20 system.

\section*{S4.~Case Studies}
\label{sec:s4-cases}

We present qualitative case studies across all modalities to illustrate how our unified
multimodal detector reasons about real, fully synthetic, and tampered media. The examples
cover challenging boundary cases and highlight the model's strengths in fine-grained spatial
localization, temporal interval detection, and detailed natural-language explanations. Across
modalities, our method consistently identifies subtle inconsistencies such as texture
misalignment, unnatural temporal dynamics, or cross-modal desynchronization while avoiding
false alarms on high-quality real content.

For images and videos, our detector produces accurate bounding boxes on small manipulated
regions and explains the visual cues behind each decision. Audio and AV-talking-head cases demonstrate the model’s
ability to detect synthetic speech artifacts, temporal editing, and audio–visual mismatch. These
examples show that the model not only outputs correct labels but also provides grounded,
interpretable reasoning aligned with human perception. Such qualitative evidence complements our
quantitative results and demonstrates the robustness and transparency of our unified approach.

In addition, we also present representative samples from \textsc{Omni-Fake-Set} and \textsc{Omni-Fake-OOD} to illustrate the visual and distributional diversity of the benchmark, as well as the high quality of the underlying data.

\begin{figure*}[t]
  \centering
  \includegraphics[width=\textwidth]{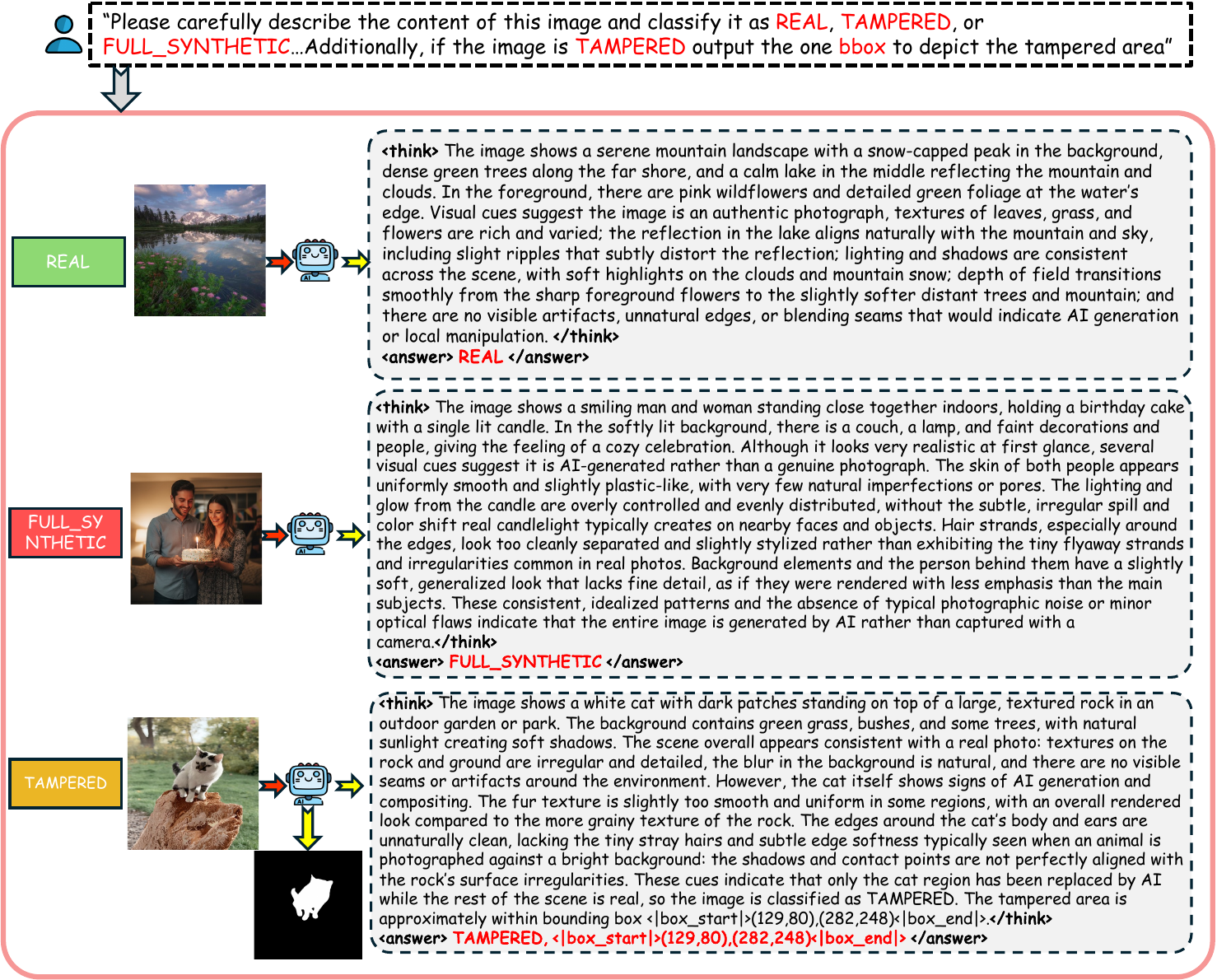}
  \caption{Image case studies with REAL, TAMPERED, and FULLY SYNTHETIC examples, including predictions, localization for TAMPERED, and explanations.}
  \label{fig:s4-image-cases}
\end{figure*}

\begin{figure*}[t]
  \centering
  \includegraphics[width=\textwidth]{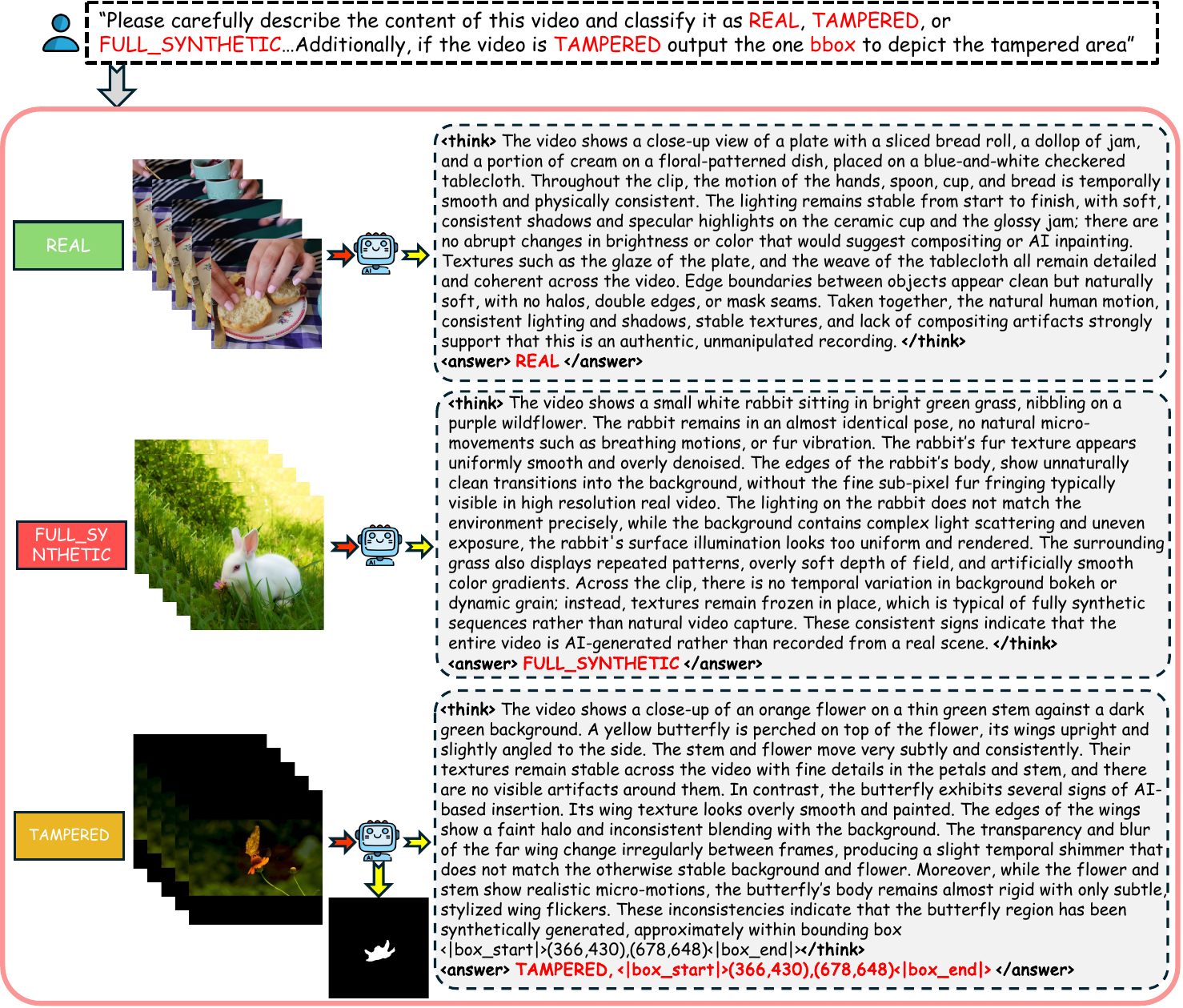}
  \caption{Video case studies with REAL, TAMPERED, and FULLY SYNTHETIC videos, showing key frames, predictions, and tampered-region localization for TAMPERED.}
  \label{fig:s4-video-cases}
\end{figure*}

\begin{figure*}[t]
  \centering
  \includegraphics[width=\textwidth]{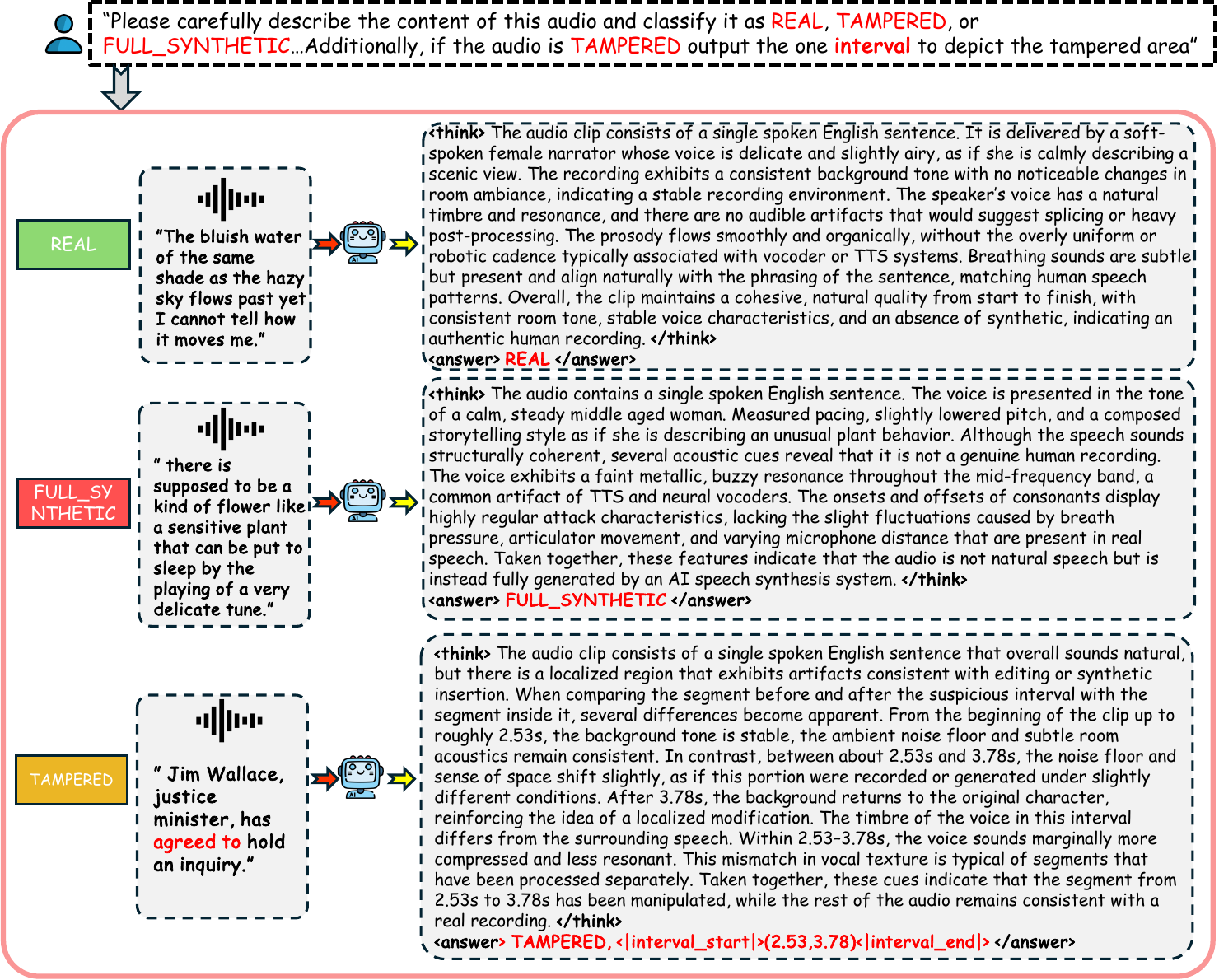}
  \caption{Audio case studies with REAL, TAMPERED, and FULLY SYNTHETIC examples. Forged temporal intervals are highlighted for TAMPERED audio, together with predictions and explanations.}
  \label{fig:s4-audio-cases}
\end{figure*}

\begin{figure*}[t]
  \centering
  \includegraphics[width=\textwidth]{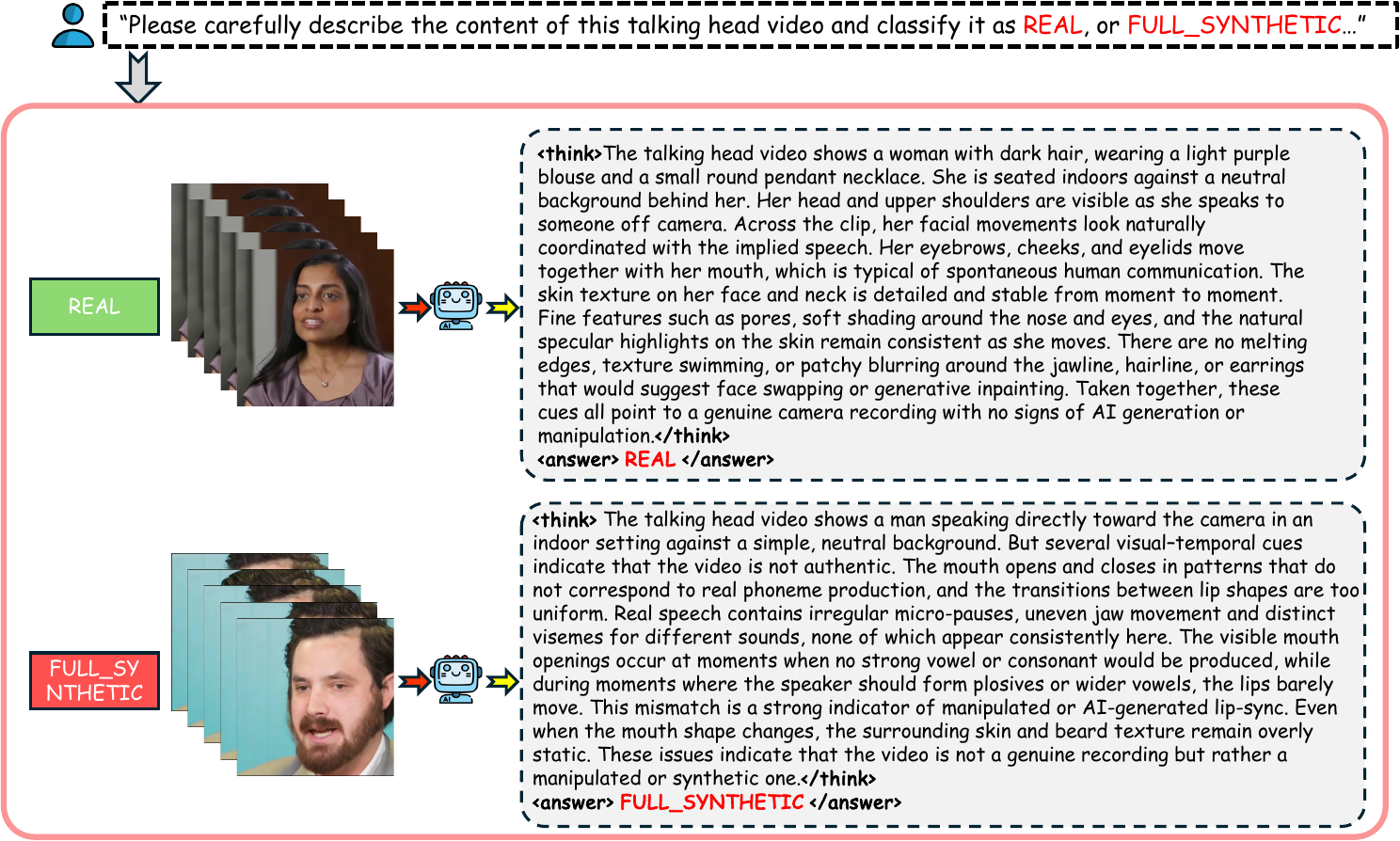}
  \caption{Audio--visual talking-head case studies with REAL and FULLY SYNTHETIC clips, showing model predictions and explanations based on audio--visual consistency.}
  \label{fig:s4-avth-cases}
\end{figure*}

\begin{figure*}[t]
  \centering
  \includegraphics[width=\textwidth]{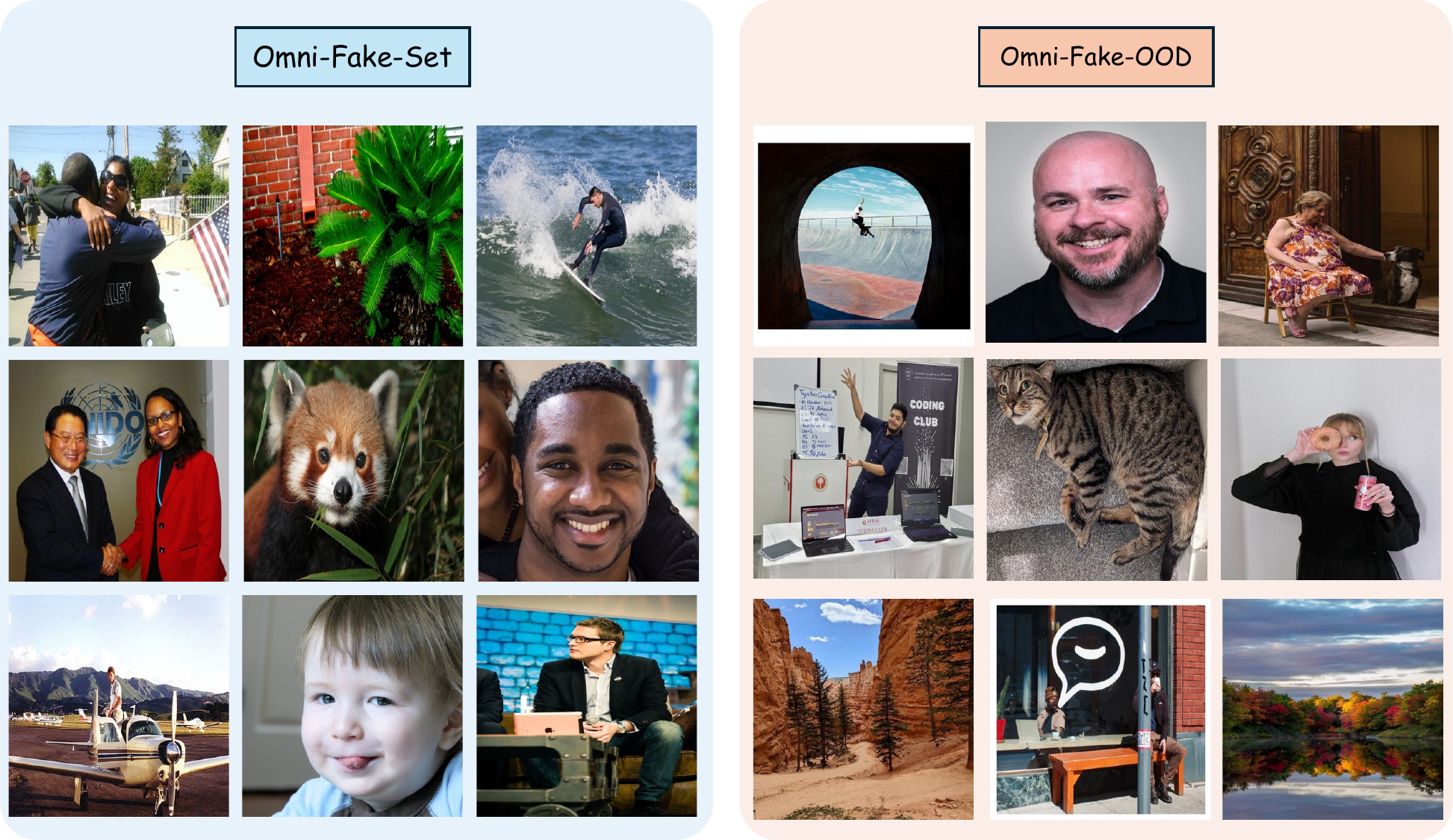}
  \caption{Representative REAL sample from the \textsc{Omni-Fake} dataset.}
  \label{fig:s4-real}
\end{figure*}

\begin{figure*}[t]
  \centering
  \includegraphics[width=\textwidth]{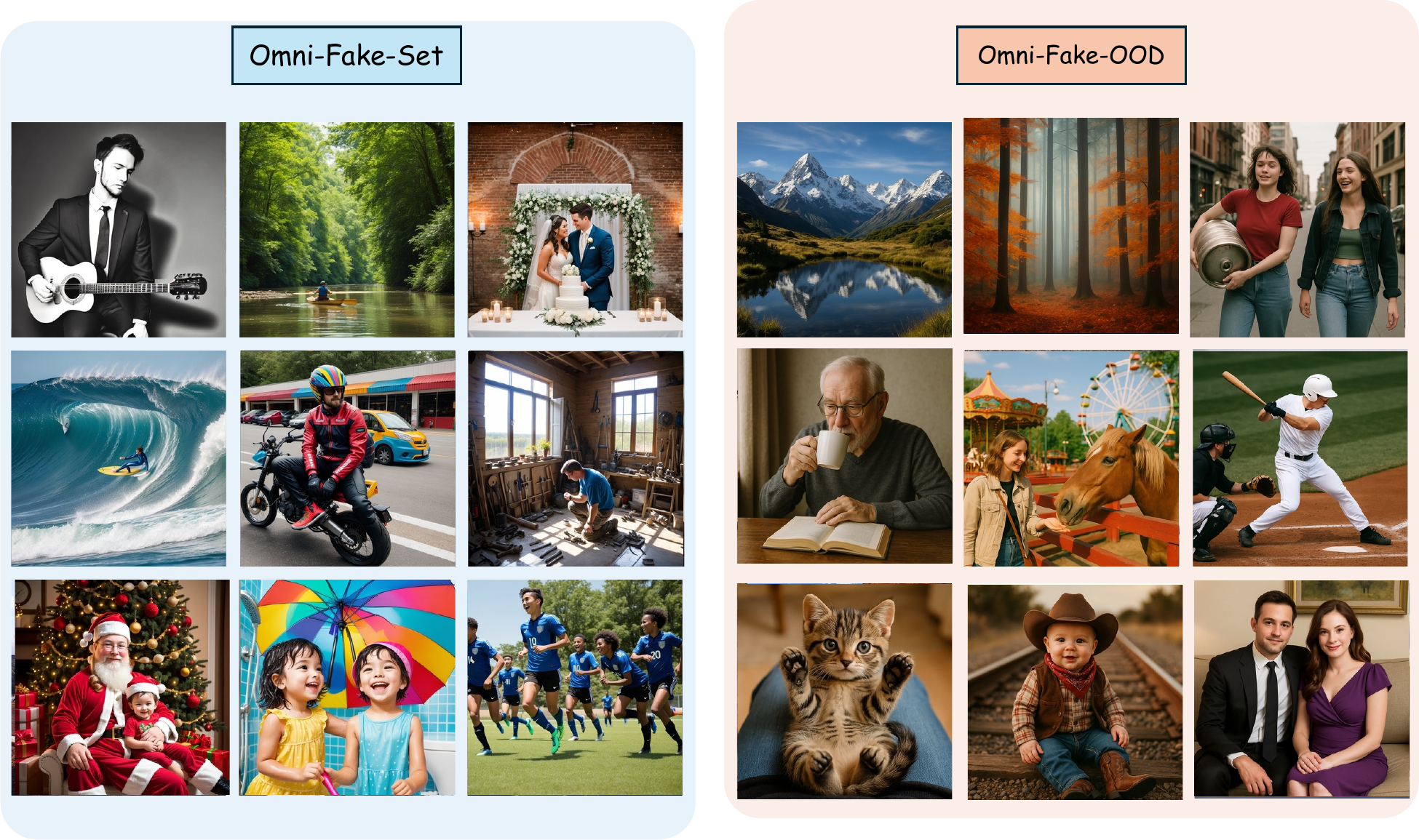}
  \caption{Representative FULL\_SYNTHETIC sample generated by modern diffusion-based models in \textsc{Omni-Fake}.}
  \label{fig:s4-syn}
\end{figure*}

\begin{figure*}[t]
  \centering
  \includegraphics[width=\textwidth]{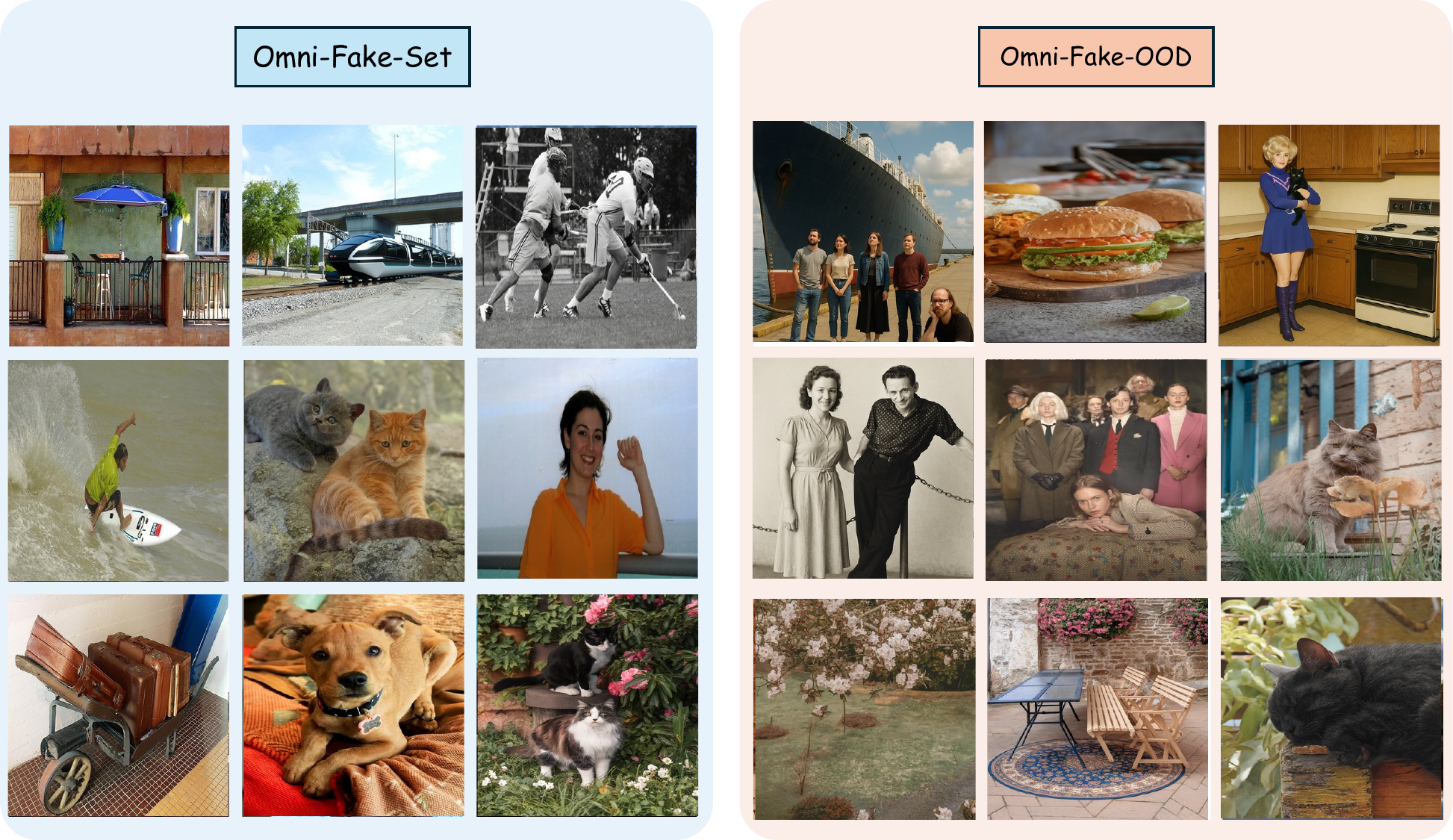}
  \caption{Representative TAMPERED sample containing localized manipulations in \textsc{Omni-Fake}.}
  \label{fig:s4-tampered}
\end{figure*}

\begin{figure*}[t]
  \centering
  \includegraphics[width=\textwidth]{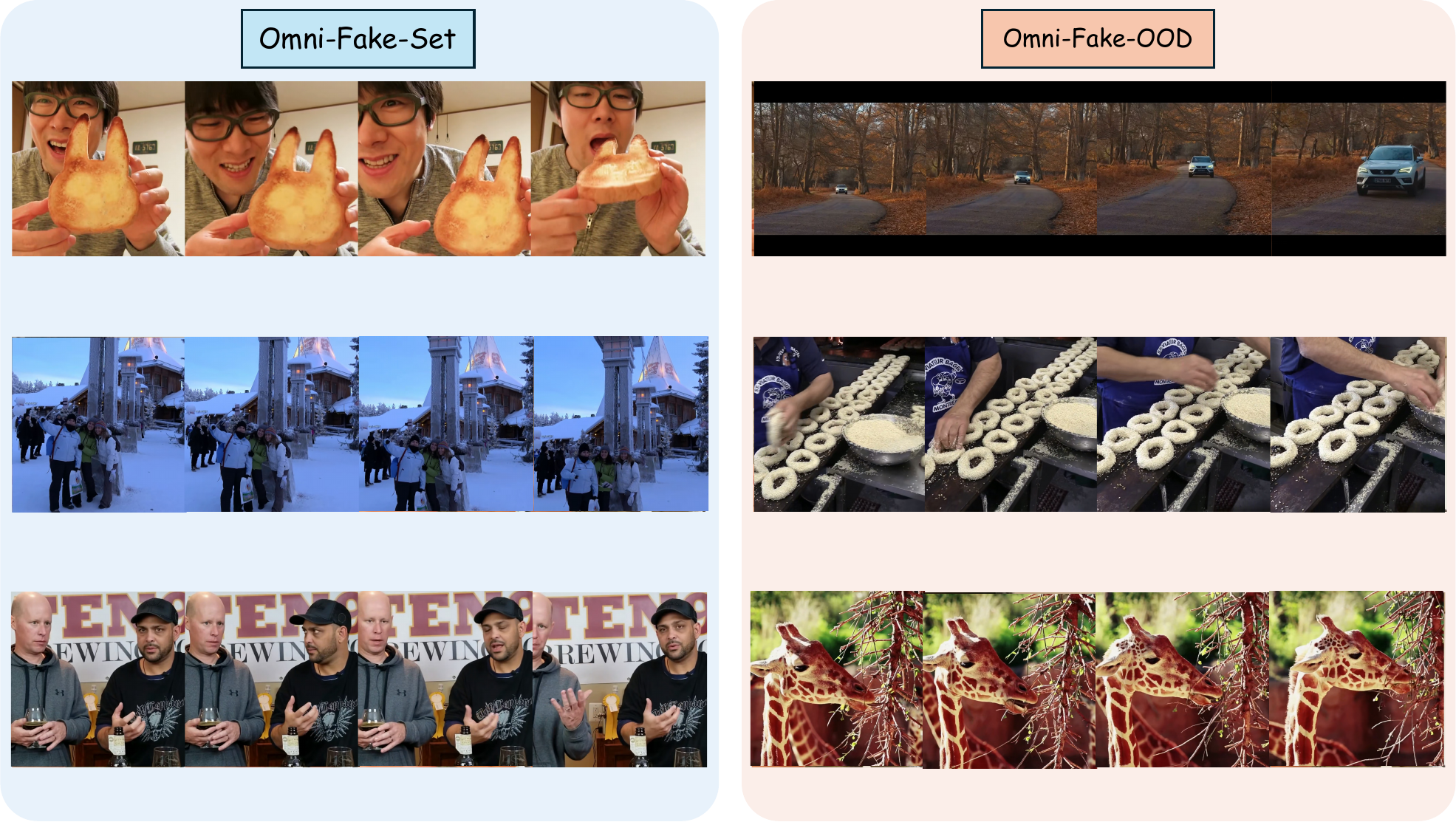}
  \caption{Representative REAL video sample from \textsc{Omni-Fake}. 
  The frames illustrate high-quality authentic motion and natural temporal dynamics.}
  \label{fig:s4-video-real}
\end{figure*}

\begin{figure*}[t]
  \centering
  \includegraphics[width=\textwidth]{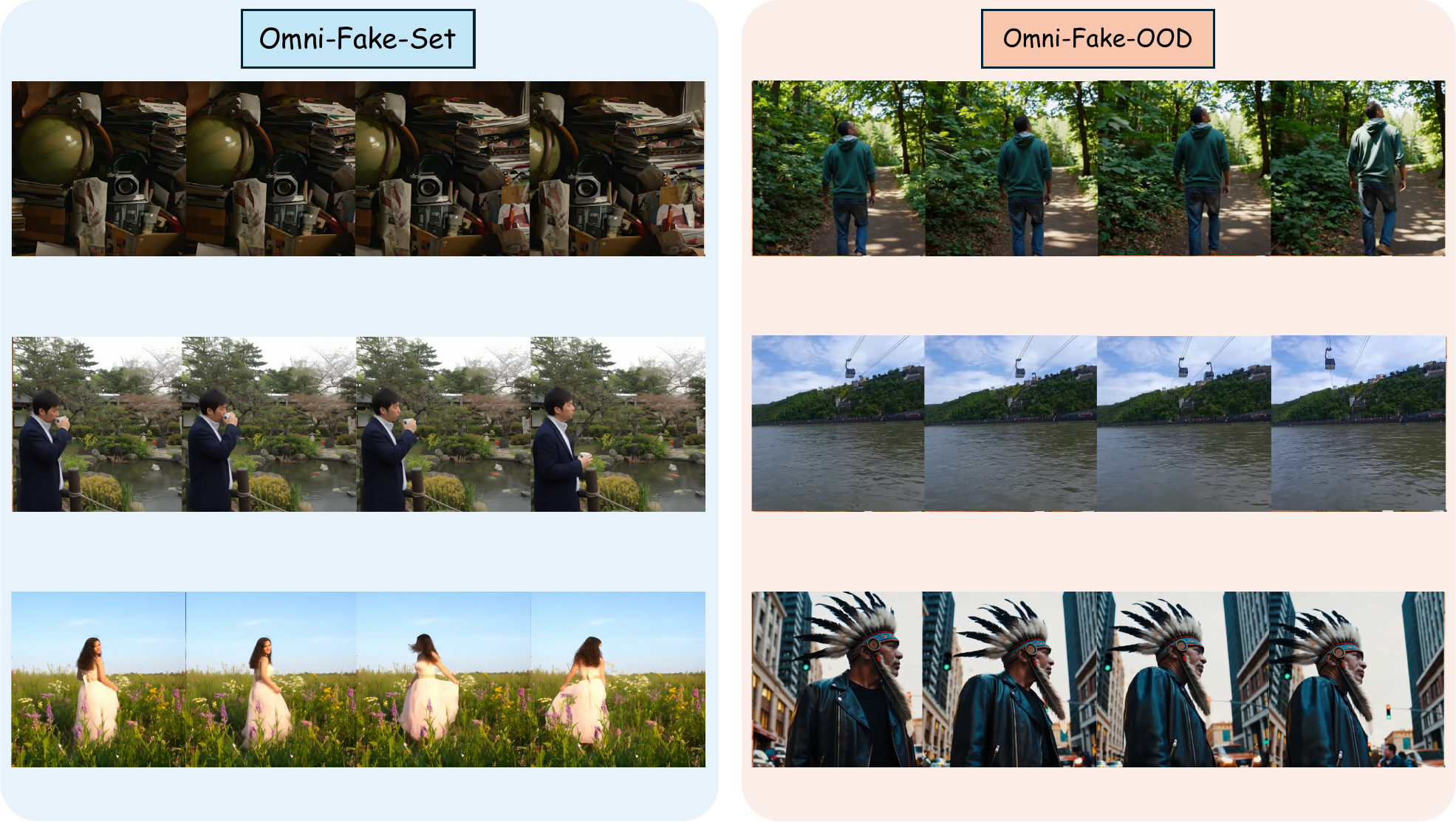}
  \caption{Representative FULLY SYNTHETIC video sample from \textsc{Omni-Fake}. 
  This example reflects typical AI-generated motion patterns and texture consistency.}
  \label{fig:s4-video-syn}
\end{figure*}

\begin{figure*}[t]
  \centering
  \includegraphics[width=\textwidth]{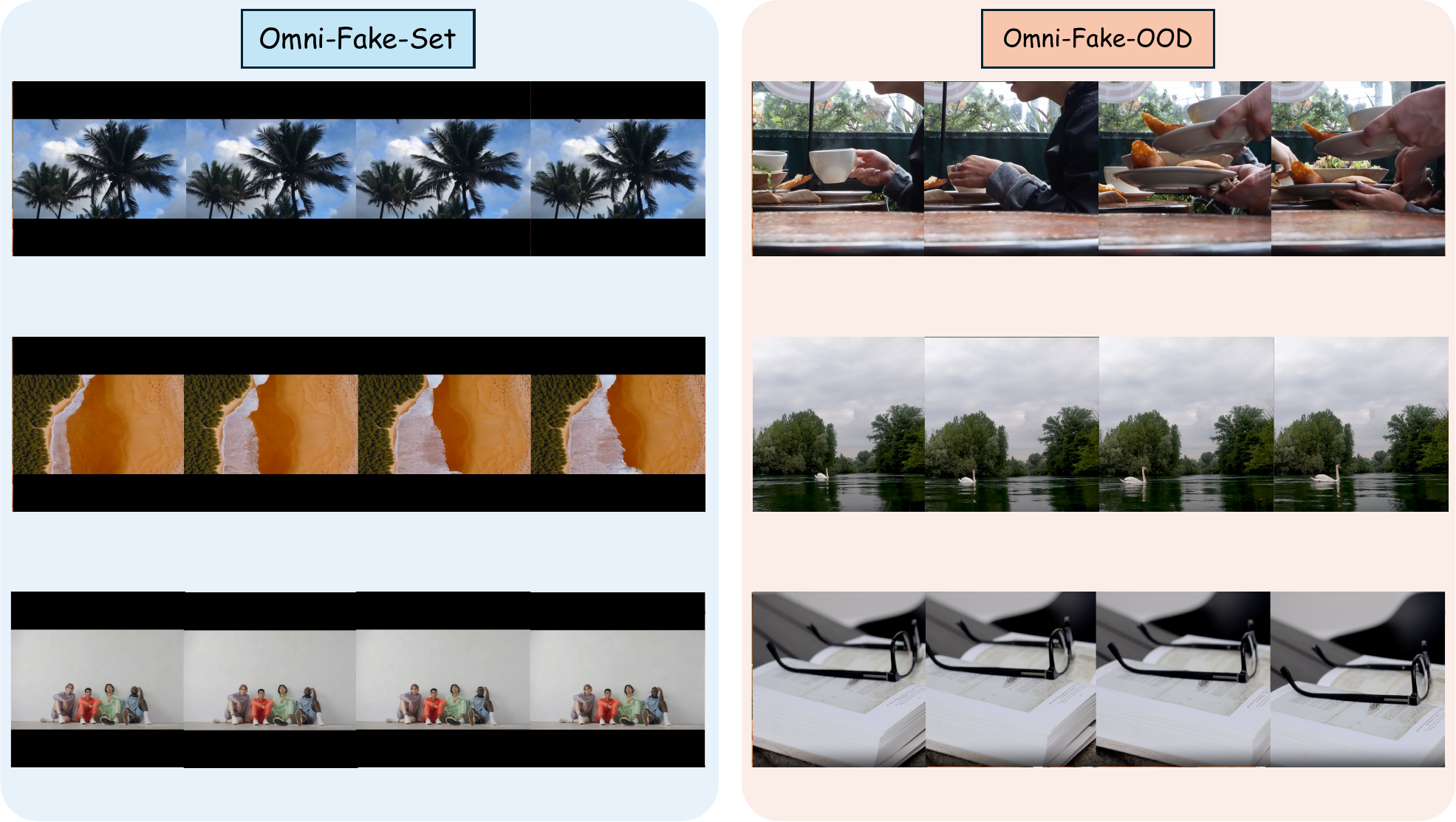}
  \caption{Representative TAMPERED video sample from \textsc{Omni-Fake}. 
  Only part of the temporal sequence is manipulated while the rest remains authentic.}
  \label{fig:s4-video-tamp}
\end{figure*}

\begin{figure*}[t]
  \centering
  \includegraphics[width=\textwidth]{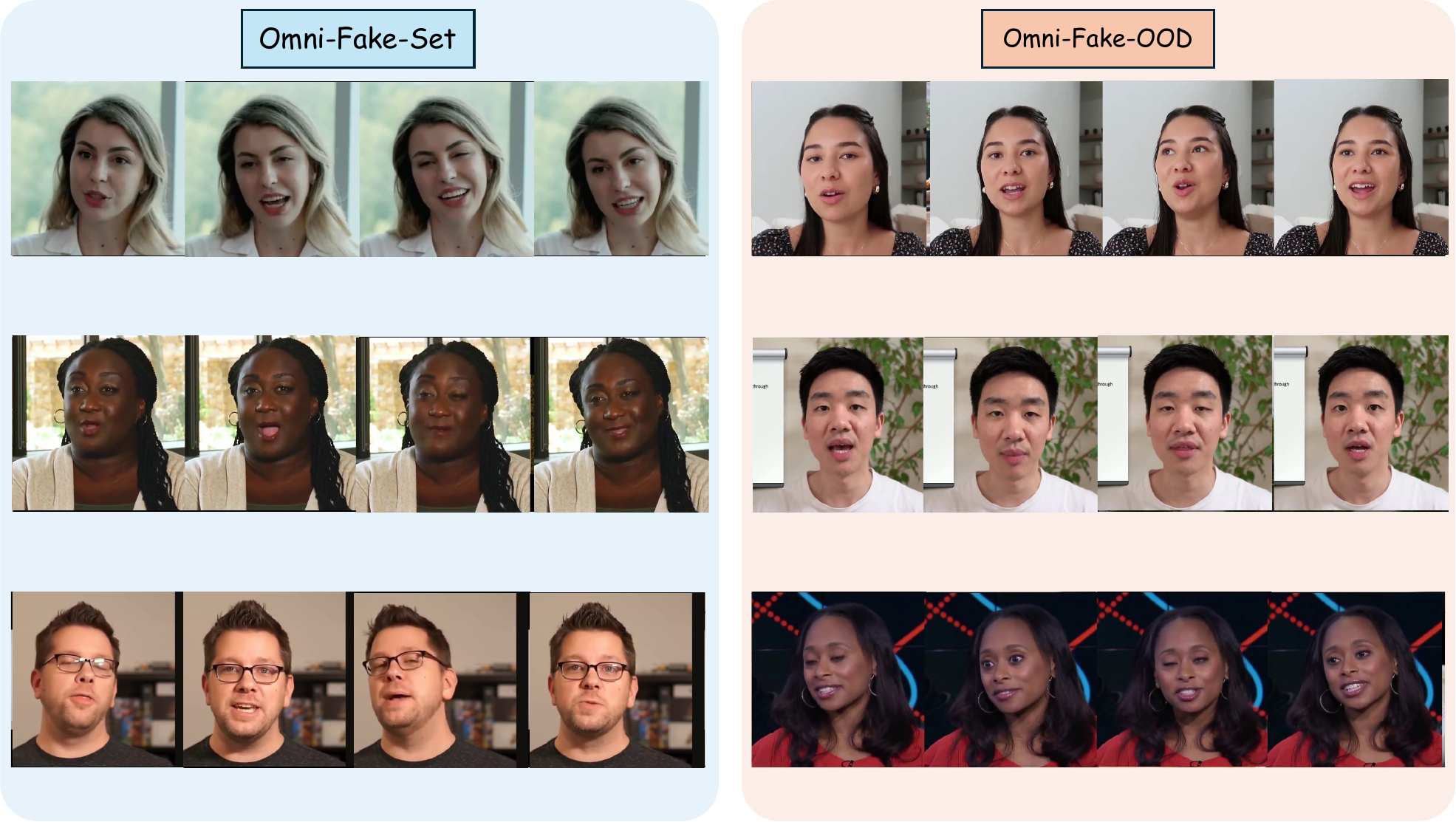}
  \caption{Representative \textbf{REAL} audio--visual talking-head samples from \textsc{Omni-Fake-Set} (left) 
  and \textsc{Omni-Fake-OOD} (right). Samples show high visual clarity, diverse recording environments, 
  and consistent lip–audio synchronization.}
  \label{fig:s3-avth-real}
\end{figure*}

\begin{figure*}[t]
  \centering
  \includegraphics[width=\textwidth]{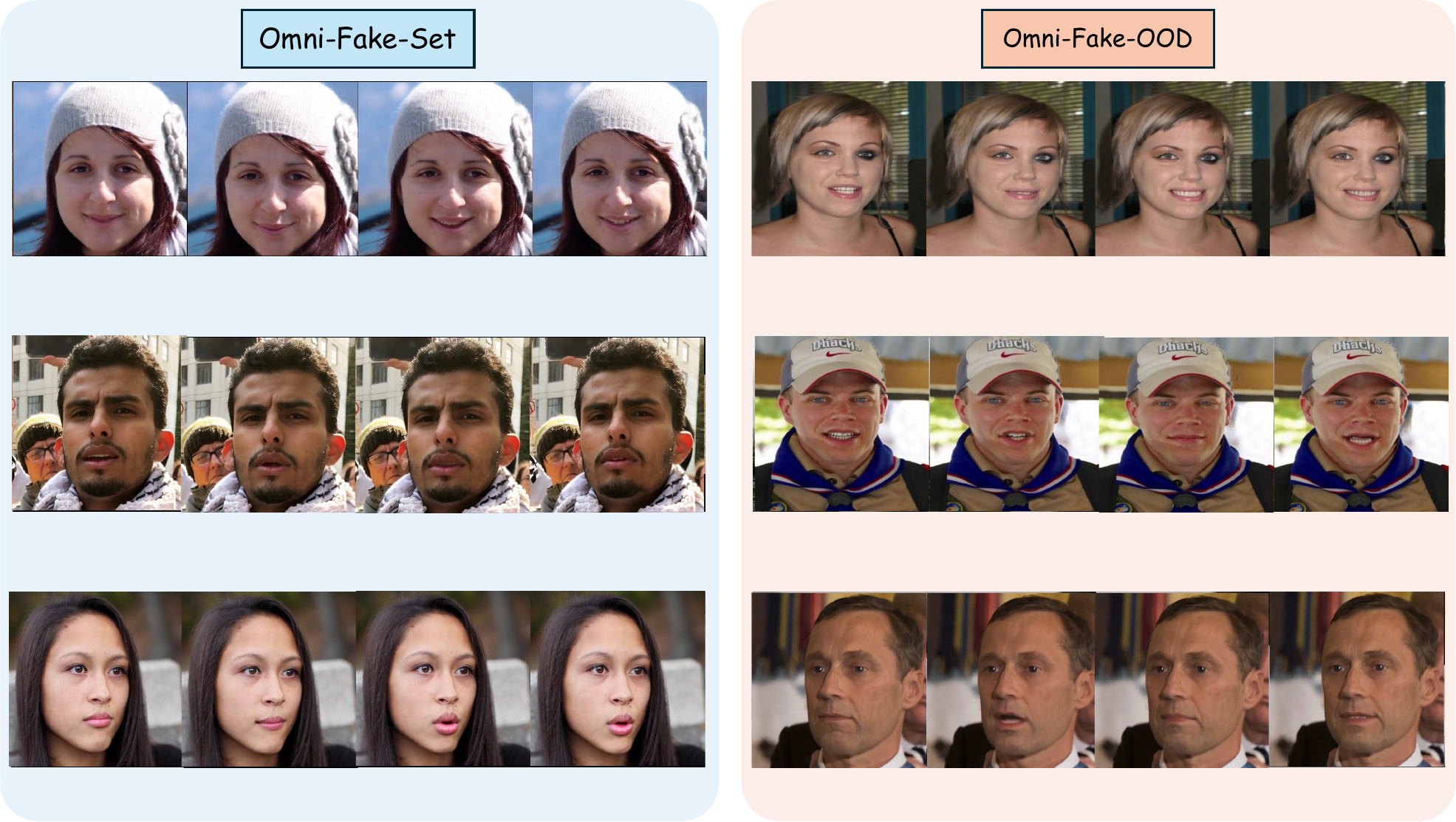}
  \caption{Representative \textbf{FULLY SYNTHETIC} audio--visual talking-head samples from \textsc{Omni-Fake-Set} (left) 
  and \textsc{Omni-Fake-OOD} (right). Synthetic samples exhibit high realism across identity appearance.}
  \label{fig:s3-avth-syn}
\end{figure*}

\noindent
\textbf{Limitation and future work.} While Omni-Fake covers four major modalities, it does not yet include some emerging formats such as 3D avatars or multilingual speech synthesis, which may become increasingly relevant as generative models advance. In addition, although the benchmark incorporates diverse manipulation types, the landscape of generative technologies evolves rapidly, and newly emerging manipulation styles may still fall outside its current scope. We view these points as natural directions for future expansion to keep Omni-Fake aligned with the growing diversity of real-world multimodal deepfakes.


\end{document}